\documentclass{article}
\usepackage{changes}

    \PassOptionsToPackage{numbers, compress}{natbib}

    \usepackage[preprint]{neurips_data_2024}

\usepackage[utf8]{inputenc} 
\usepackage[T1]{fontenc}    
\usepackage{hyperref}       
\usepackage{url}            
\usepackage{booktabs}       
\usepackage{amsfonts}       
\usepackage{nicefrac}       
\usepackage{microtype}      
\usepackage{xcolor}         
\usepackage{graphicx}
\usepackage{tablefootnote}
\usepackage{caption}
\usepackage{subcaption}
\usepackage[most]{tcolorbox}
\usepackage{tabularx,colortbl,xcolor}

\usepackage{booktabs}
\usepackage{amsmath}
\usepackage{mathtools} 

\usepackage{booktabs}
\usepackage{multirow}
\usepackage{amsfonts}
\usepackage{amssymb}
\usepackage{tabulary}
\usepackage{enumitem}
\usepackage{adjustbox}
\usepackage{makecell}
\usepackage{tcolorbox}
\usepackage{multirow, multicol}
\usepackage{pifont}
\usepackage{booktabs}
\usepackage{tablefootnote}
\usepackage{arydshln}
\usepackage{xstring}
\newcommand{\xmark}{\ding{55}}%

\usepackage{titletoc}
\usepackage{tocloft}

\title{TrustSQL: Benchmarking Text-to-SQL Reliability with Penalty-Based Scoring}

\author{
    Gyubok Lee\quad
    Woosog Chay\quad
    Seonhee Cho\quad
    Edward Choi \\
    KAIST AI\\
    \texttt{\{gyubok.lee, benchay, seonhee.cho, edwardchoi\}@kaist.ac.kr} \\
    }

\begin{document}

\maketitle

\begin{abstract}
Text-to-SQL enables users to interact with databases using natural language, simplifying the retrieval and synthesis of information.
Despite the remarkable success of large language models (LLMs) in translating natural language questions into SQL queries, widespread deployment remains limited due to two primary challenges.
First, the effective use of text-to-SQL models depends on users' understanding of the model's capabilities—the scope of questions the model can correctly answer.
Second, the absence of abstention mechanisms can lead to incorrect SQL generation going unnoticed, thereby undermining trust in the model's output.
To enable wider deployment, it is crucial to address these challenges in model design and enhance model evaluation to build trust in the model's output.
To this end, we introduce \textit{TrustSQL}, a novel comprehensive benchmark designed to evaluate ~\textit{text-to-SQL reliability}—defined as a model's ability to correctly handle any type of input question by generating correct SQL queries for feasible questions and abstaining from generating infeasible ones (\textit{e.g.}, due to schema incompatibility or functionalities beyond SQL).
We evaluate existing methods using a novel penalty-based scoring metric with two modeling approaches: (1) pipeline-based methods combining SQL generators with infeasible question detectors and SQL error detectors for abstention; and (2) unified methods using a single model for the entire task.
Our experimental results reveal that achieving high scores under severe penalties requires significant effort and provide a new perspective on developing text-to-SQL models for safer deployment. TrustSQL is available at \url{https://github.com/glee4810/TrustSQL}.
\end{abstract}

\section{Introduction}

Text-to-SQL is a task that involves generating SQL queries from natural language (NL) questions, a challenge that has persisted for several decades~\cite{shwartz2022tabular}.
This task is crucial both within and beyond the realm of machine learning and natural language processing (NLP), due to the ubiquity of databases in various fields such as medicine, finance, retail, and telecommunications.
Accessing structured data stored in these databases through natural language can unlock the true potential of big data, enabling domain experts to uncover new quantitative insights~\cite{van2024tabular}.

\begin{figure}[t!]
    \centering
    \begin{subfigure}[t!]{0.32\textwidth}
        \centering
        \includegraphics[width=\textwidth]{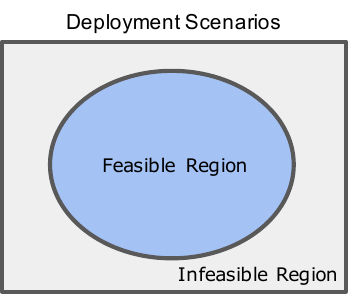}
        \caption{}
        \label{fig:sub2}
    \end{subfigure}
    \hfill
    \begin{subfigure}[t!]{0.32\textwidth}
        \centering
        \includegraphics[width=\textwidth]{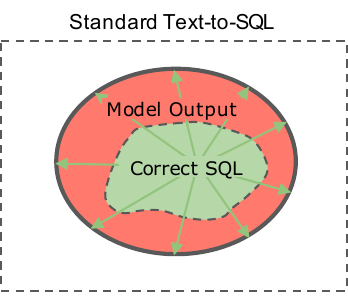}
        \caption{}
        \label{fig:sub1}
    \end{subfigure}
    \hfill
    \begin{subfigure}[t!]{0.32\textwidth}
        \centering
        \includegraphics[width=\textwidth]{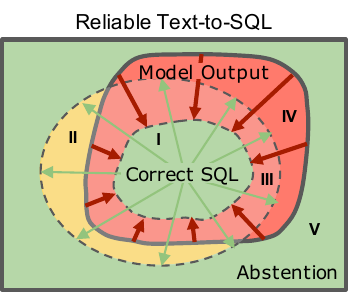}
        \caption{}
        \label{fig:sub3}
    \end{subfigure}
    \caption{
\textbf{(a) Deployment Scenarios} include both feasible and infeasible user questions regarding text-to-SQL tasks (\textit{e.g.}, requests outside the database schema or SQL functionalities).
\textbf{(b) Standard Text-to-SQL} aims to maximize correct SQL query generation. This focus can result in incorrect model outputs going unnoticed, increasing risk in deployment. The green and red regions represent correct and incorrect SQL generations, respectively. To effectively utilize a text-to-SQL model, users must be familiar with its capabilities, which vary by model—the types of questions each model can correctly answer—setting a high barrier for potential users.
\textbf{(c) Reliable Text-to-SQL} removes this barrier and aims to enhance trust in the model by minimizing incorrect outputs \textbf{(III)} and \textbf{(IV)} through abstention (red arrow), while maximizing correct outputs \textbf{(I)} within the feasible region (green arrow). By shrinking the red regions \textbf{(III)} and \textbf{(IV)}, and subsequently the yellow region \textbf{(II)}, users will begin to trust the model's outputs. Each region is further explained in the footnote\protect\footnote{}.}
\label{fig:task_definition}
\end{figure}

To effectively utilize these models, users need to be familiar with their capabilities. These include the types of questions each individual model can accurately answer, the information it has access to, and what operations can or cannot be performed via SQL. However, many potential users who could greatly benefit from the model are unfamiliar with these capabilities and may find it challenging to use, as their questions may frequently fall outside the model’s scope (beyond the correct SQL in Figure~\ref{fig:sub1}). The challenge intensifies in real-world deployment scenarios where user questions include not only those feasible for the model but also infeasible ones, such as requests requiring knowledge beyond the database content or SQL functionalities (as shown in Figure~\ref{fig:sub2}). Moreover, the absence of abstention mechanisms in current text-to-SQL modeling can lead to incorrect model outputs for both feasible and infeasible questions going unnoticed, potentially outweighing the benefits of model use. If a model, by design, could selectively answer correct SQL queries while abstaining from answering the rest, it would significantly boost user confidence, thereby enhancing trust in the model and encouraging its use~\cite{whitehead2022reliable, ren2022out}.

\footnotetext{The five key regions in Figure~\ref{fig:sub3} are: \textbf{(I)} correct SQL generation for feasible questions (desirable for text-to-SQL reliability), \textbf{(II)} abstention for feasible questions (neither strictly desirable nor undesirable, thus neutral), \textbf{(III)} incorrect SQL generation for feasible questions (undesirable), \textbf{(IV)} incorrect SQL generation for infeasible questions (undesirable), and \textbf{(V)} correct abstention for infeasible questions (desirable). All five distinct cases are all incorporated into our penalty-based scoring metric introduced in Section~\ref{sec:evaluation_metric}, and their actual examples are provided in Supplementary~\ref{appendix:rs_qualitative}.}

To address the aforementioned challenges, we present \textit{TrustSQL}, a new benchmark designed to assess text-to-SQL reliability.
We define \textit{text-to-SQL reliability} as a model’s ability to handle any input questions correctly through accurate SQL generation and abstention, thereby reducing the red and yellow regions by covering the entire region in green, as shown in Figure~\ref{fig:sub3}).
Unlike standard text-to-SQL tasks defined within the feasible region (Figure~\ref{fig:sub1}), real user questions span both feasible and infeasible task regions, including requests beyond the database content and SQL functionalities  (Figure~\ref{fig:sub2}).
We constructed this new comprehensive setting using three domain-specific datasets—ATIS~\cite{atis}, Advising~\cite{advising}, and EHRSQL~\cite{lee2022ehrsql}—along with one cross-domain dataset—Spider~\cite{yu2018spider}.
In constructing TrustSQL, we reflected common error cases undermining text-to-SQL reliability through careful data splitting and incorporation of various infeasible questions, similar to red teaming.
We then evaluated existing methods using a novel penalty-based scoring metric designed specifically to measure text-to-SQL reliability (details in Section~\ref{sec:evaluation_metric}).
Our experimental results reveal that while most advanced methods seem to perform well in penalty-less settings, achieving high scores under severe penalty settings remains highly challenging.
From the high penalty perspective, knowing when to abstain precisely carries more weight than rashly attempting to generate SQL queries, underscoring the need for substantial efforts towards gaining trust in model outputs.


\section{Dataset Construction}

In this section, we begin by identifying common error cases that undermine text-to-SQL reliability during deployment, causing answers to reside in the red regions in Figure~\ref{fig:sub3}. Then, we introduce our strategies to incorporate each case into the dataset construction process. Below are the common error cases in text-to-SQL deployment:

\vspace{-2mm}
\begin{itemize}[leftmargin=5mm] 
  \item \textbf{Complexity of SQL generation}: User questions, especially those posed by domain experts, are often too complex for models to answer accurately. This complexity arises from the inherent difficulty of the queries~\cite{yu2018spider, li2023can, hazoom2021sede, lee2022ehrsql}, as in Table~\ref{tab:feasible_sample} in Supplementary~\ref{appendix:data_complexity}, or from the model encountering unfamiliar phrases and entities specific to the database~\cite{suhr-etal-2020-exploring, lee-etal-2021-kaggledbqa}. Models that generate SQL without the ability to abstain from such cases risk compromising trust in their output. We include these dimensions of query difficulty \textbf{[P1]} and familiarity \textbf{[P2]} in feasible data construction.
  \item \textbf{Ambiguity in user queries}: User questions are often ambiguous, particularly when they refer to columns or values (\textit{i.e.}, database records) without precise specification. Resolving the user's true intent involves multi-turn interactions for clarification, which is beyond the scope of this work.
  \item \textbf{Schema incompatibility}: Users who are unfamiliar with the exact database schema may inadvertently ask questions that cannot be processed because the information they seek does not exist in the existing columns. Previous works construct such cases by deleting and perturbing existing columns~\cite{triagesql, known_dte}. Instead, we include three distinct types within this case \textbf{[P3]} to provide more granular views on model behavior.
  \item \textbf{Scope of SQL functionalities}: Users may not always be familiar with the scope of SQL functionalities, leading to requests that require complex numerical operations, data analysis, or scheduling appointments, which are beyond standard SQL execution. We include such questions \textbf{[P4]} in infeasible data construction.
  \item \textbf{Need for external knowledge}: Users may ask questions that involve factual world knowledge that goes far beyond the database content, either fully (\textit{e.g.}, `What airport is in Wilder, KY?') or partially (\textit{e.g.}, `Find me patients born after WWII'). We consider the former cases \textbf{[P5]} when constructing infeasible questions, while excluding the latter. 
\end{itemize}

\begin{table}[t!]
\caption{Sample infeasible questions for each type for the orchestra database in Spider, containing 4 tables with 23 columns. Samples for domain-specific databases are not shown here for brevity. `Note' indicates hypothetical columns in existing tables (\textit{i.e.}, columns not present in the database we use for text-to-SQL) or a task name used for question annotation.}
\centering
\renewcommand{\arraystretch}{1.2}
\begin{adjustbox}{width=\columnwidth,center}  
\begin{tabular}{clc}
\toprule
\textbf{Type} & \multicolumn{1}{c}{\textbf{Question}} & \textbf{Note} \\
\midrule
column-surface	& What is the average attendance capacity for all the shows? & show.attendance\_capacity \\
\midrule
column-related	& Which orchestra has the most members?	& orchestra.number\_of\_member \\
\midrule
column-unrelated & What aircraft does Charles Dutoit own? & conductor.aircraft \\
\midrule
non-sql	& \makecell[l]{Identify the top three significant features that influence attendance.} & feature importance \\
\midrule
ext-know &  What advantages does diversity bring to an orchestra? & - \\
\bottomrule
\end{tabular}
\end{adjustbox}
\label{tab:infeasible_sample}
\end{table}


\subsection{Feasible Data Construction}
\label{sec:feasible_question}

\paragraph{Dataset selection} To establish the benchmark data covering comprehensive settings, we selected three complex domain-specific text-to-SQL datasets, namely, ATIS (flight information)~\cite{atis}, Advising (education)~\cite{advising}, and EHRSQL (healthcare)~\citep{lee2022ehrsql}.
Additionally, we included one cross-domain dataset, Spider.
The distinction between domain-specific and cross-domain text-to-SQL settings, along with the complexity of each dataset, is detailed in Supplementary~\ref{appendix:data_complexity}.

\vspace{-2mm}
\paragraph{Data preprocessing} Our preliminary data analysis revealed that ATIS and Advising require significant preprocessing due to issues such as numerous query duplications, inconsistencies, and poor-quality natural language (NL) questions (each issue discussed in Supplementary~\ref{appendix:data_preprocessing}).
We implemented several strategies to correct these issues, including creating new question paraphrases for all input NL questions using GPT-4, followed by a manual review by the authors.
For further details, please refer to Supplementary~\ref{appendix:data_preprocessing}).
The databases of the four datasets and their corresponding question-SQL pairs, after preprocessing, are included as questions in the feasible task region (light blue region in Figure~\ref{fig:sub2}).
To reflect the query difficulty \textbf{[P1]}, we adopted the SQL hardness criteria following~\cite{pourreza2023din-sql} across all datasets.
Detailed statistics for each hardness level are provided in Table~\ref{tab:feasible_sample} in Supplementary~\ref{appendix:data_complexity}.
Query familiarity \textbf{[P2]} is later introduced in the data splitting process~(\S\ref{sec:data_splitting}).

\begin{table}[t!]
\centering
\scriptsize
\renewcommand{\arraystretch}{1.0}
\caption{Dataset statistics of TrustSQL. TrustSQL includes an equal number of unanswerable and answerable questions in the test sets. For $Q$, the three numbers, separated by '/', denote the size of training, validation, and test sets.}
\begin{tabular}{lcccc}
\hline
& ATIS & Advising & EHRQSL & Spider \\
\hline
All Questions ($Q \cup Q^{c}$) & 2555 & 2769 & 7473 & 8561 \\ 
\hline
\quad Feasible Question ($Q$)& 1114 / 489 / 476 & 1170 / 533 / 533 & 4674 / 931 / 934  & 7000 / 507 / 527 \\
\hline
\quad Infeasible Question ($Q^{c}$) & 476 & 533 & 934  & 527\\
\quad $^{\llcorner}$ \texttt{non-sql}   & 99 & 110 & 191 & 100 \\
\quad $^{\llcorner}$ \texttt{surface}   & 93 & 102 & 185 & 100 \\
\quad $^{\llcorner}$ \texttt{unrelated} & 94 & 110 & 187 & 100 \\
\quad $^{\llcorner}$ \texttt{related}   & 98 & 107 & 189 & 100 \\
\quad $^{\llcorner}$ \texttt{ext-know}  & 92 & 104 & 182 & 127 \\
\hline
\end{tabular}
\label{tab:data_statistic}
\end{table}

\subsection{Infeasible Data Construction}
\label{sec:infeasible_question}

Text-to-SQL models deployed in real-world scenarios often encounter user requests that fall into the infeasible task region (gray region in Figure~\ref{fig:sub2}) due to going beyond the database content or SQL functionalities, and thus do not have corresponding SQL queries.
To evaluate models in a more realistic environment, we manually annotated NL questions by incorporating three common infeasible error cases as follows:

\vspace{-2mm}
\paragraph{Schema incompatibility} \textbf{[P3]}: Instead of randomly dropping existing columns in the database~\cite{triagesql, known_dte}, we introduce the concept of \textit{hypothetical columns} as target columns for infeasible questions. Those questions are asked as if their corresponding SQL queries exist, but in fact do not, since no such columns exist in the database. We categorized these into three types: \texttt{column-surface}, \texttt{column-related}, and \texttt{column-unrelated}, representing attacks based on surface-form and semantic column similarities and dissimilarities, respectively.

\vspace{-3mm}
\begin{itemize}[itemsep=-2pt, leftmargin=3.5mm] 
  \item \texttt{column-surface}: Questions referring to hypothetical columns that resemble existing columns in different tables, or have similar surface forms (\textit{e.g.}, a hypothetical `age' column in the `Professionals' table that resembles the `age' column in the `Dogs' table in the `dog\_kennels' database).
  \item \texttt{column-related}: Questions referring to hypothetical columns that could naturally integrate into the existing database schema (\textit{e.g.}, a hypothetical column named `duration' in the `song' table). This question type can be seen as a manual version of the ADD perturbations in ADVETA~\cite{adveta}.
  \item \texttt{column-unrelated}: Questions referring to hypothetical columns that are randomly sampled from other databases, which are mostly unnatural (\textit{e.g.}, a hypothetical `product\_category' column in the `battles' database).
\end{itemize}

\vspace{-3mm}
\paragraph{Beyond SQL functionalities} \textbf{[P4]}: To capture diverse user requests that exceed SQL execution (\texttt{non-sql}), we compiled a list of NLP and data analytics tasks and used GPT-4 to identify what could be done as an AI assistant if any tools can be used given the database content. Based on a list of 27 tasks (\textit{e.g.}, anomaly detection, visualization, and web search), we manually annotated NL questions relevant to or referring to existing columns in each database (thus, context-driven annotation), simulating users expecting the model to perform tasks not solely addressed through SQL.  

\vspace{-2mm}
\paragraph{Need for external knowledge} \textbf{[P5]}: To construct NL questions involving external knowledge (\texttt{ext-know}), we prompted GPT-4 to generate questions that require domain expertise, qualitative insights, or causal understanding. We continued this process until we achieved a balanced set of samples comparable to other types of infeasible questions.

After manual annotation of infeasible NL questions, the authors used GPT-4 to iteratively review them to ensure that they were truly infeasible.
The sample annotated questions are provided in Table~\ref{tab:infeasible_sample}.

\subsection{Data Splitting}
\label{sec:data_splitting}

The core principle for data splitting is to create varying query familiarity \textbf{[P2]} for feasible data and introduce the unexpectedness of infeasible data for model evaluation. Table~\ref{tab:data_statistic} summarizes the overall data statistics in TrustSQL.

\vspace{-3mm}
\paragraph{Feasible data} We define `seen SQL' and `unseen SQL' to introduce query familiarity \textbf{[P2]} (only applicable within feasible data).
Samples are considered `seen SQL'  when their corresponding SQL structures\footnote{A SQL structure refers to a specific combination of SQL components with specific tables and columns, excluding actual data values.} are present in the training set but have different paraphrased NL questions in the validation and test sets.
`Unseen SQL' samples have SQL structures not present in the training set. 
We use these different query familiarities \textbf{[P2]} included in the validation and test sets to analyze model performances. 
The data splitting process for domain-specific datasets is as follows: 1) Split feasible data into training, validation, and test sets in a 6:2:2 ratio based on NL questions (same SQL structures across splits, but different NL questions); 2) Remove 20\% of non-overlapping SQL structures from the validation and test sets; 3) Remove the total 40\% of these SQL structures from the training set, resulting in 60\% `seen SQL' in all splits, while validation and test sets contain 80\% SQL structures (60\% `seen SQL' and 20\% `unseen SQL'). For Spider, we divide the development split in half for new validation and test sets (10 databases each), all `unseen SQL,' with no overlaps with the training set.


\vspace{-3mm}
\paragraph{Infeasible data} We add the annotated infeasible NL questions of five types (\texttt{column-surface}, \texttt{column-related}, \texttt{column-unrelated}, \texttt{non-sql}, and \texttt{ext-know}) all to the test set in balanced proportions to reflect their unexpected encounter in deployment.


\section{Benchmarking Text-to-SQL Reliability}

\subsection{Task}
\label{sec:task}

Reliable text-to-SQL aims to develop models that only generate correct SQL queries while abstaining in any other cases, including both feasible ($\mathcal{Q}$) and infeasible questions ($\mathcal{Q}^{c}$), to enhance text-to-SQL reliability.
Optimized under this principle, users can effectively use these models without having extensive knowledge about the models' capabilities.
They can also be less concerned about the model producing incorrect outputs, as it would abstain from answering questions beyond its capabilities.
This increased reliability will lead to wider adoption of text-to-SQL models, empowering domain experts to utilize natural language interfaces to access and analyze the vast amounts of data stored in databases.

\vspace{-2mm}
\subsection{Evaluation Metric}
\label{sec:evaluation_metric}

In evaluating reliable text-to-SQL performance, we categorize the model's outputs into three distinct categories: desirable (green), undesirable (red), and neutral (yellow), as illustrated in Figure~\ref{fig:sub3}.
These categories reflect different types of consequences brought about by the model's outputs.
Desirable outputs, providing \textit{utility}, include correct SQL generation \textbf{(I)} and abstentions \textbf{(V)}. 
However, the severity of the consequences of undesirable outputs can vary based on the context of the deployment.
In less critical settings (\textit{e.g.}, experimental scenarios), we might be lenient towards these undesirable outputs and focus solely on the model's utility.
In contrast, in high-stakes environments (\textit{e.g.}, hospitals), undesirable outputs can have severe repercussions, potentially outweighing any utility provided by the model.
Therefore, we define the relative magnitude of the negative impact compared to the positive impact as \textit{risk}.
This risk includes incorrect SQL generations \textbf{(III)} and abstentions \textbf{(IV)}, weighted by a penalty factor, $c$, namely the ratio of the negative impact to the positive impact.
Neutral (yellow) cases \textbf{(II)} are neither strictly desirable nor strictly undesirable; therefore, we consider them neither utility nor risk.
Examples of instances for each case are provided in Appendix~\ref{appendix:rs_qualitative}.

In summary, text-to-SQL reliability can be quantified as the difference between utility and risk. To formalize this notion mathematically, we define this penalty-based scoring for text-to-SQL—\textit{reliability score (RS)}—as follows:

\begin{equation}
\small
\phi_{c}(x) = 
    \begin{cases}
      1 & \text{if $x \in \mathcal{Q}$; $g(x) = 1$; $Acc(x) = 1$,} \\
      0 & \text{if $x \in \mathcal{Q}$; $g(x) = 0$,} \\
      -c & \text{if $x \in \mathcal{Q}$; $g(x) = 1$; $Acc(x) = 0$,} \\
      -c & \text{if $x \in \mathcal{Q}^{c}$; $g(x) = 1$,} \\
      1 & \text{if $x \in \mathcal{Q}^{c}$; $g(x) = 0$,} \\
    \end{cases}
\label{eq:reliability-metric}
\end{equation}

where $Acc(x)$ is the execution accuracy in text-to-SQL and $g(x) = 1$ and $g(x) = 0$ indicate the model's decision to answer or abstain, respectively. 

For each sample, a score of 1 is assigned when the SQL is generated correctly for $x \in \mathcal{Q}$ \textbf{(I)} or when the model abstains for $x \in \mathcal{Q}^{c}$ \textbf{(V)}, indicating utility.
A score of $0$ is given when the model decides not to generate an SQL for $x \in \mathcal{Q}$ \textbf{(II)}, indicating no risk and no utility.
A penalty of $-c$ is assigned when the generated SQL is incorrect \textbf{(III)} or when the model attempts to generate an SQL for $x \in \mathcal{Q}^{c}$ \textbf{(IV)}, indicating risk.
Actual examples of each case are provided in Supplementary~\ref{appendix:rs_qualitative}.
After calculating each sample-level score, the aggregate RS can be computed by taking the mean, $\frac{1}{N}\sum_{x}\phi_{c}(x)$, resulting in $\phi_{c}$.


In our experiments, we consider three options for the penalty value $c$: 0, 10, and $N$, where $N$ represents the total number of samples to be evaluated.

\vspace{-1mm}
\begin{itemize}[leftmargin=4.5mm, itemsep=-1pt, topsep=-0.5pt]
    \item \(c = 0\): The most lenient scenario. This does not penalize any mistakes made by the model. If we only handle feasible questions, this score is equivalent to the standard text-to-SQL accuracy.
    \item \(c = 10\): A moderately strict scenario. Every 10 correct answers weigh the same as one incorrect answer.
    \item \(c = N\): The most strict scenario. A single mistake, even if all other answers were correct, leads to a negative score.
\end{itemize}

An RS($c$) of 100\% indicates perfect text-to-SQL reliability, meaning the model can handle all inputs correctly (\textit{i.e.}, covering all the regions with green in Figure~\ref{fig:sub3}). RS($c$) > 0 means the utility is greater than the risk, making it deployable in the context of text-to-SQL reliability.


\section{Experiments}
\vspace{-2mm}
\subsection{Modeling}
\label{sec:modeling}

In this section, we describe various methods for developing models to enhance text-to-SQL reliability. We categorize the methods into two broad modeling approaches: pipeline-based and unified. For each approach, we develop both open-source and commercial API-based models, as these choices are often utilized one or the other depending on various factors in deployment, such as budget constraints, technical expertise, and privacy concerns.

\begin{table*}[t!]
\caption{Model comparison in TrustSQL involving both feasible and infeasible questions evaluated in $\phi_{_{0}}$, $\phi_{_{10}}$, and $\phi_{_{N}}$. AT, AD, EH, and SP refer to ATIS, Advising, EHRSQL, and Spider, respectively. \textsc{Abstain-All} blindly abstains from predicting all input questions, reflecting the ratio of infeasible questions. All in \% for readability.}
\centering
\renewcommand{\arraystretch}{1.2}
\begin{adjustbox}{width=\textwidth,center}
\large
\begin{tabular}{ccccccccccccccc}
\hline


\multicolumn{3}{c}{}
& \multicolumn{4}{c}{$\phi_{0}$}
& \multicolumn{4}{c}{$\phi_{10}$}
& \multicolumn{4}{c}{$\phi_{N}$} \\ 

\cmidrule(lr){4-7}
\cmidrule(lr){8-11}
\cmidrule(lr){12-15}

& &
& AT & AD & EH & SP
& AT & AD & EH & SP
& AT & AD & EH & SP \\ 


\hline

\multicolumn{2}{c}{} & \textsc{Abstain-All}
& 50.0 & 50.0 & 50.0 & 50.0 & 50.0 & 50.0 & 50.0 & 50.0 & 50.0 & 50.0 & 50.0 & 50.0 \\

\hline
\hline
\cellcolor{white!50} \multirow{7}{*}{Open}

& \multirow{3}{*}{Pipeline}
& \textsc{CLS$_M$} → \textsc{T5}
& 55.7 & 48.1 & 62.7 & \underline{61.1} & -160.7 & -154.5 & -60.4 & -304.2 & -20.5K & -21.6K & -22.9K & -38.4K\\

& & \textsc{T5} → \textsc{Error$_{M}$} 
& \textbf{73.0} & 73.8 & 75.5 & 60.3 & -59.3 & -133.5 & -115.6 & -121.8 & -12.5K & -22.0K & -35.6K & -19.1K
 \\

& & \textsc{CLS$_M$} → \textsc{T5} → \textsc{Error$_M$}
& 64.3 & 58.9 & 66.5 & \textbf{64.3} & \underline{13.9} & -21.8 & 14.6 & -66.6 & \underline{-4.7K} & -8.5K & -9.6K & -13.7K
 \\

\cdashline{2-15}

& \multirow{4}{*}{Unified}
& \textsc{T5[MaxEnt]}
& 63.8 & \textbf{74.6} & \textbf{85.3} & 58.3 & 6.0 & \textbf{28.6} & \underline{36.1} & -71.7 & -5.4K & \textbf{-4.8K} & \underline{-9.1K} & -13.6K  \\

& & \textsc{T5[MaxProb]}
& 61.6 & \underline{74.4} & \textbf{85.3} & 56.7 & \textbf{15.3} & \underline{18.1} & \textbf{36.6} & \underline{-18.2} & \textbf{-4.3K} & \underline{-5.9K} & \textbf{-9.0K} & \underline{-7.8K} \\

& & \textsc{T5[FeatMD]}
& 67.5 & 57.1 & 67.6 & 50.1 & -9.1 & -315.3 & 14.1 & \textbf{50.1} & -7.2K & -39.6K & -9.9K & \textbf{50.1}
 \\

& & \textsc{T5[FeatRMD]}
& \underline{72.8} & 55.8 & 76.4 & 49.9 & -48.0 & -352.3 & -153.2 & -296.4 & -11.4K & -43.4K & -42.8K & -36.5K
 \\

\hline
\hline

\cellcolor{white!50} \multirow{7}{*}{API}
& \cellcolor{white!50} \multirow{5}{*}{Pipeline}
& \textsc{CLS$_P$} → \textsc{SQLPrompt}
& \textbf{88.9} & \textbf{88.7} & \textbf{85.3} & 67.1 & -2.5 & -3.2 & -48.0 & -126.5 & -8.6K & -9.7K & -24.8K & -20.3K \\

\cellcolor{white!50} 
& \cellcolor{white!50}  & \textsc{CLS$_P$ → DIN-SQL}
& 61.4 & \underline{75.7} & 61.5 & \textbf{75.0} & -304.1 & -146.6 & -310.6 & \underline{-39.8} & -34.7K & -23.6K & -69.4K & -12.0K
 \\

& & \textsc{CLS$_{P}$} → \textsc{MAC-SQL}
& 73.2 & 71.0 & 66.1 & \underline{74.4} & -169.4 & -198.2 & -260.0 & -46.1 & -23.0K & -28.6K & -60.8K & -12.6K \\

& & \textsc{SQLPrompt} → \textsc{Error$_P$}
& 52.4 & 55.8 & 54.3 & 72.1 & 24.1 & 5.2 & 9.4 & -107.2 & -2.6K & -5.3K & -8.3K & -18.8K \\

& & \textsc{CLS$_P$} → \textsc{SQLPrompt} → \textsc{Error$_P$}
& 54.5 & 58.9 & 57.4 & 67.6 & \textbf{51.4} & \textbf{45.8} & \textbf{46.7} & -41.5 & \textbf{-245.5} & \textbf{-1.3K} & \textbf{-1.9K} & \underline{-11.4K} \\
\cdashline{2-15}

& \multirow{2}{*}{Unified}
& \textsc{SQLPrompt[Demo]}
& \underline{73.4} & 70.4 & 68.6 & 66.8 & -190.2 & -226.1 & -241.3 & -265.3 & -25.0K & -31.5K & -57.8K & -34.9K
 \\

& & \textsc{SQLPrompt[Voting]}
& 66.2 & 67.4 & \underline{73.4} & 50.7 & \underline{41.0} & \underline{40.2} & \underline{45.0} & \textbf{29.8} & \underline{-2.3K} & \underline{-2.8K} & \underline{-5.2K} & \textbf{-2.1K}\\


\hline

\end{tabular}
\end{adjustbox}

\label{tab:main_all}
\end{table*}

\vspace{-2mm}
\subsubsection{Pipeline-based Approach}
\label{sec:pipeline-based}

In conventional NLP task formulations, enhancing text-to-SQL reliability can be addressed by combining three different tasks: infeasible question detection~\cite{triagesql, known_dte}, SQL generation~\cite{pourreza2023din-sql, wang2023mac}, and SQL error detection~\cite{chen2023teaching}.

\vspace{-2mm}
\paragraph{Open-source} For SQL generation, we use \textsc{T5} (\texttt{T5-3B}), feeding questions followed by a serialized schema listing all tables and columns~\cite{suhr-etal-2020-exploring}. To incorporate abstention mechanisms, we use SQLCoder (\texttt{sqlcoder-7b-2}) for infeasible question detection (before SQL generation) and error detection (after SQL generation). For infeasible question detection, we fine-tune SQLCoder on TriageSQL~\cite{triagesql}, denoted as \textsc{CLS$_{M}$}. For error detection, we train SQLCoder based on the generated SQLs and their sample-level accuracy from trained T5 on the TrustSQL validation set plus samples of infeasible questions in TriageSQL, denoted as \textsc{Error$_{M}$}. Details about training these models are provided in Supplementary~\ref{appendix:modeling_detail_opensource}.
When denoting the combined methods, we use arrows to indicate data flow between models, such as \textsc{CLS$_{M}$} → \textsc{T5} indicating that question detection comes first, followed by SQL generation.


\vspace{-1mm}
\paragraph{API-based} For SQL generation, we employ three methods using GPT-3.5 (\texttt{gpt-3.5-turbo-0125}): \textsc{SQLPrompt}, SQL generation with few-shot prompting~\cite{chang2023prompt}, \textsc{DIN-SQL}~\cite{pourreza2023din-sql}, a state-of-the-art task decomposition framework in Spider, and \textsc{MAC-SQL}~\cite{wang2023mac}, a state-of-the-art multi-agent framework in BIRD~\cite{li2023can}. 
We adjust text prompts for \textsc{DIN-SQL} and \textsc{MAC-SQL} for domain-specific datasets. 
For abstention, we use GPT-4o for both infeasible question detection and error detection. We provide the modeling details in Supplementary~\ref{appendix:modeling_detail_api}.


\vspace{-1mm}
\subsubsection{Unified Approach}
\label{sec:unified}

To achieve text-to-SQL reliability using a single model, we explore various text-to-SQL prompts that incorporate abstention mechanisms. Additionally, we utilize multiple uncertainty estimation methods, leveraging SQL generators’ hidden representations and probabilities, to enable abstention.

\vspace{-2.0mm}
\paragraph{Open-source} For SQL generation, we use T5 (\texttt{T5-3B}) due to its wide use in selective generation~\cite{ren2022out} and calibration~\cite{stengel2023calibrated} in NLP. For abstention, we leverage the hidden representations of the input and the output probabilities by adopting uncertainty estimation to detect samples that are out-of-distribution (OOD) for abstention. Below are the methods we use:

\vspace{-2.0mm}
\begin{itemize}[leftmargin=3.5mm, itemsep=-1pt, topsep=-0.5pt] 
  \item Entropy-based: Each sample is assigned the maximum entropy value among the generated SQL tokens, denoted as \textsc{MaxEnt}~\citep{lee2022ehrsql}.
  \item Probability-based: Each sample is assigned the minimum of the highest token probabilities from the token sequence, denoted as \textsc{MaxProb}~\citep{stengel2023calibrated}.
  \item Distance-based: Mahalanobis distance calculation based on the distribution of the training data representations, denoted as \textsc{FeatMD}~\citep{lee2018simple, ren2022out}, and relative Mahalanobis distance calculation with a background distribution using the TriageSQL dataset, denoted as \textsc{FeatRMD}~\citep{ren2021simple, ren2022out}.
\end{itemize}

These methods require calibrating thresholds to draw a decision boundary between in-distribution and out-of-distribution. Our proposed heuristic method is provided in Supplementary~\ref{appendix:thresholding}.

\begin{table*}[t!]
\caption{Model comparison in TrustSQL using only feasible data. In this table, $\phi_{_{0}}$ is equivalent to execution accuracy in standard text-to-SQL. The scores in shaded cells represent SQL performance without abstention.}
\centering
\renewcommand{\arraystretch}{1.2}
\begin{adjustbox}{width=\textwidth,center}
\large
\begin{tabular}{ccccccccccccccc}
\hline


\multicolumn{3}{c}{}
& \multicolumn{4}{c}{$\phi_{0}$}
& \multicolumn{4}{c}{$\phi_{10}$}
& \multicolumn{4}{c}{$\phi_{N}$} \\ 

\cmidrule(lr){4-7}
\cmidrule(lr){8-11}
\cmidrule(lr){12-15}

& &
& AT & AD & EH & SP
& AT & AD & EH & SP
& AT & AD & EH & SP \\ 


\hline
\rowcolor{lightgray!50}
\cellcolor{white!50} \multirow{8}{*}{Open}
& \multirow{4}{*}{Pipeline}
\cellcolor{white!50}&  \textsc{T5}
& 57.8 & 72.6 & 86.8 & 50.9 & -364.5 & -201.3 & -44.9 & -440.6 & -20.0K & -14.5K & -12.2K & -25.8K \\
& &\textsc{CLS$_M$} → \textsc{T5}
& 32.6 & 28.3 & 42.6 & \textbf{49.0} & -188.0 & -56.1 & -32.3 & -414.0 & -10.5K & -4.5K & -7.0K & -24.4K \\

& & \textsc{T5} → \textsc{Error$_{M}$} 
& \textbf{57.6} & \textbf{70.5} & \underline{79.0} & \underline{41.4} & -91.6 & -115.2 & -22.7 & -116.1 & -7.0K & -9.8K & -9.4K & -8.3K \\

& & \textsc{CLS$_M$} → \textsc{T5} → \textsc{Error$_M$}
& 32.4 & 28.3 & 38.2 & 39.8 & \underline{-30.7} & -28.0 & -14.2 & -110.1 & \underline{-3.0K} & \underline{-3.0K} & -4.9K & -7.9K
 \\


\cdashline{2-15}

& \multirow{4}{*}{Unified}
& \textsc{T5[MaxEnt]}
& 31.7 & 52.7 & 77.6 & 32.1 & -41.8 & \textbf{-3.6} & \underline{48.7} & -72.3 & -3.5K & \textbf{-2.9K} & \underline{-2.6K} & -5.5K
 \\ 

& & \textsc{T5[MaxProb]}
& 27.7 & 53.7 & 77.6 & 22.2 & \textbf{-18.5} & \underline{-10.1} & \textbf{49.8} & \underline{-40.4} & \textbf{-2.2K} & -3.3K & \textbf{-2.5K} & \underline{-3.3K}

 \\

& & \textsc{T5[FeatMD]}
& 35.5 & 64.7 & 41.4 & 0.2 & -113.7 & -175.4 & -3.5 & \textbf{0.2} & -7.1K & -12.7K & -4.2K & \textbf{0.2}

 \\

& & \textsc{T5[FeatRMD]}
& \underline{45.8} & \underline{67.7} & \textbf{86.5} & 37.4 & -193.7 & -187.4 & -36.6 & -279.5 & -11.4K & -13.5K & -11.4K & -16.7K

 \\

\hline
\hline

\rowcolor{lightgray!50}
\cellcolor{white!50} \multirow{10}{*}{API}
& \cellcolor{white!50} \multirow{8}{*}{Pipeline}
& \textsc{SQLPrompt}
& 81.5 & 82.2 & 82.3 & 54.5 & -103.4 & -96.1 & -93.3 & -400.9 & -8.7K & -9.4K & -16.3K & -23.9K
\\
\rowcolor{lightgray!50}
\cellcolor{white!50} & \cellcolor{white!50} &\textsc{DIN-SQL}
& 26.5 & 56.7 & 34.9 & 75.0 & -708.8 & -376.6 & -616.1 & -175.5 & -35.0K & -23.0K & -60.8K & -13.1K \\
\rowcolor{lightgray!50}
\cellcolor{white!50} & \cellcolor{white!50} &\textsc{MAC-SQL}
& 49.2 & 46.3 & 44.4 & 73.6 & -448.7 & -490.2 & -511.2 & -190.1 & -23.7K & -28.6K & -51.9K & -13.8K
 \\

& &\textsc{CLS$_P$} → \textsc{SQLPrompt}
& \underline{79.0} & \underline{80.1} & \underline{81.8} & 38.0 & -91.2 & -77.5 & -73.4 & -311.2 & -8.0K & -8.3K & -14.4K & -18.4K
 \\

& & \textsc{CLS$_P$ → DIN-SQL}
& 24.2 & 54.0 & 34.4 & \underline{55.8} & -694.3 & -364.4 & -595.2 & \underline{-115.0} & -34.2K & -22.2K & -58.8K & \underline{-8.9K}

 \\

& & \textsc{CLS$_{P}$} → \textsc{MAC-SQL}
& 47.7 & 44.7 & 43.4 & 54.5 & -425 & -467.5 & -496.3 & -129.6 & -22.5K & -27.3K & -50.4K & -9.6K \\

& & \textsc{SQLPrompt} → \textsc{Error$_P$}
& 9.9 & 19.3 & 16.2 & 50.7 & \underline{3.6} & -5.1 & 1.2 & -243.5 & \underline{-290.1} & \underline{-1.3K} & \underline{-1.4K} & -15.4K \\

& & \textsc{CLS$_P$} → \textsc{SQLPrompt} → \textsc{Error$_P$}
& 9.0 & 18.2 & 16.1 & 35.5 & 2.7 & \underline{-4.3} & \underline{7.5} & -180.8 & \underline{-290.1} & \textbf{-1.2K} & \textbf{-783.9} & -11.4K \\
\cdashline{2-15}

& \multirow{2}{*}{Unified}
& \textsc{SQLPrompt[Demo]}
& \textbf{79.4} & \textbf{81.1} & \textbf{82.2} & \textbf{74.6} & -122.3 & -108.4 & -88.0 & -179.7 & -9.5K & -10.0K & -15.8K & -13.3K
 \\

& & \textsc{SQLPrompt[Voting]}
& 34.5 & 37.9 & 49.6 & 4.7 & \textbf{5.0} & \textbf{13.5} & \textbf{20.7} & \textbf{-2.8} & \textbf{-1.4K} & \underline{-1.3K} & -2.7K & \textbf{-395.3} \\


\hline

\end{tabular}
\end{adjustbox}

\label{tab:main_feasible}
\end{table*}

\vspace{-0.2mm}
\paragraph{API-based} We use two prompting methods with GPT-3.5, as explained below:

\begin{itemize}[leftmargin=3.5mm, itemsep=-1pt, topsep=-0.5pt] 
  \item \textsc{SQLPrompt[DEMO]}: Few-shot demonstrations of similar feasible questions with their corresponding SQL queries and similar infeasible questions, each labeled with `not\_answerable'~\cite{sun2021conditionalqa}. The order of demonstrations is randomly chosen.
  \item \textsc{SQLPrompt[Voting]}: Using the same prompts as \textsc{SQLPrompt}, we sample the answers five times~\citep{wang2022self} and consider the SQL generation only if all samples unanimously agree; otherwise, we abstain.
\end{itemize}

\section{Results}
\vspace{-1mm}
\subsection{Model Comparison}
\vspace{-2mm}
\paragraph{Full Data} Table~\ref{tab:main_all} shows the performance of each method in the TrustSQL benchmark. For open-source models, the probability-based (\textit{i.e.}, \textsc{T5[MaxEnt]}) and entropy-based (\textit{i.e.}, \textsc{T5[MaxProb]}) methods in the unified approach consistently outperform other methods across different penalty settings. For API-based models, \textsc{CLS$_P$} → \textsc{SQLPrompt} performs best in RS(0), while \textsc{CLS$_P$} → \textsc{SQLPrompt} → \textsc{Error$_P$} excels in stricter settings (\textit{i.e.}, RS(10) and RS(N)). This indicates that both question detection and error detection are crucial for enhancing text-to-SQL reliability. 
\textsc{SQLPrompt[Voting]} shows notable performance across all penalties, particularly given its simplicity.

\vspace{-3mm} 
\paragraph{Feasible Data Only} Evaluating models on feasible data measures unnecessary abstention and the impact of error detectors, indicated by the performance drop between the SQL performance without abstention (shaded cells) and the models with abstention in RS(0). Similar to the full data setting, open-source models, \textsc{T5[MaxEnt]} and \textsc{T5[MaxProb]}, tend to perform best with increasing penalties. 
API-based models using the unified approach, particularly \textsc{SQLPrompt[Voting]}, outperform \textsc{SQLPrompt[Demo]} as the penalty increases. High performance in RS(0) indicates frequent abstention, resulting in lower performance with higher penalties. A more detailed analysis of each component in pipeline-based methods is provided in Supplementary~\ref{appendix:error_analysis}. 



\vspace{-2mm}
\subsection{Discussion}
\vspace{-1mm}
\paragraph{Q1: Why is \textsc{All-abstain} considered deployable in text-to-SQL reliability?}
In the context of standard text-to-SQL, \textsc{All-abstain} in Table~\ref{tab:main_all} cannot be considered deployable since it contradicts the very purpose of the task. However, in the context of text-to-SQL reliability, \textsc{All-abstain} achieves an RS of 50 on all $c$ by correctly handling 50\% of all input questions (\textit{i.e.}, the proportion of infeasible questions in TrustSQL) with certainty. This approach may be more reliable (absolutely no risk in all settings) than a model optimized for standard text-to-SQL, which always generates SQL queries that may be incorrect (risk potentially higher than utility).

\vspace{-2mm}
\paragraph{Q2: What is an easy way to enhance text-to-SQL reliability?}
An easy way to enhance the RS to at least a deployable level (\textit{i.e.}, >0) in strict settings is to generate only those SQL queries for which the model has high confidence, while abstaining from generating any when uncertain. The experimental results demonstrate that this approach is particularly effective in settings where incorrect generation carries a high penalty. For instance, as shown in Table~\ref{tab:main_all}, the \textsc{T5[FeatMD]} model on Spider generates only one correct SQL query and abstains from the rest due to our strict heuristic thresholding (see Supplementary~\ref{appendix:thresholding}). Efforts to reduce risk can gain more RS scores than efforts to increase utility. However, achieving higher RS (\textit{i.e.}, >50\%) with such cautious SQL generation requires another line of effort, which has not been explored extensively.

\vspace{-2mm}
\paragraph{Q3: Why is TrustSQL necessary as a text-to-SQL benchmark?}

Existing benchmarks lack penalty-based scoring and considerations for infeasible questions. TrustSQL addresses these gaps by modeling SQL generation with abstention and handling error cases from infeasible question detection, which are then propagated to SQL generation and error detection. This comprehensive approach encourages new modeling aspects to enhance reliability. 
For instance, the entries \texttt{column-surface} and \texttt{non-sql} in Table~\ref{tab:infeasible_sample} are examples that fail to be handled using GPT-4o-based infeasible question detection and error detection methods (\textit{i.e.}, \textsc{CLS$_P$} → \textsc{SQLPrompt} → \textsc{Error$_P$}). 
Conventional task settings cannot explore such cases, showing how these models respond to infeasible questions. For instance, the query `SELECT avg(attendance) FROM show' is generated for the question `What is the average attendance capacity for all the shows?' (where only the attendance column exists, not the capacity), returning a result of 1326.4 from the database.
TrustSQL's comprehensive approach provides a new perspective on error cases in text-to-SQL generation, encouraging the development of more reliable models.



\vspace{-2mm}
\section{Related Works}
\vspace{-2mm}
\paragraph{Model Reliability in Text-to-SQL}
Text-to-SQL has attracted significant attention for its practical applications. Key single-domain datasets include ATIS~\citep{atis}, GeoQuery~\citep{geography}, Advising~\cite{advising}, SEDE~\cite{hazoom2021sede}, and EHRSQL~\cite{lee2022ehrsql}. For cross-domain tasks, Spider~\cite{yu2018spider} and BIRD~\cite{li2023can} serve as essential benchmarks for SQL generation. 
However, these benchmarks assume that all input questions should generate SQL, lacking incentives for developing models to abstain. This absence of safety measures undermines reliable deployment. Some efforts focus on detecting infeasible questions given database schemas~\cite{triagesql, known_dte}, but they do not integrate SQL generation or consider the error-handling capabilities of text-to-SQL models. Our work aims to integrate comprehensive reliability measures, aligning the benchmark more closely with model trust.


\vspace{-2mm}
\paragraph{Uncertainty Estimation in NLP} Uncertainty estimation is vital for AI reliability~\cite{hendrycks2021unsolved}. The methods include probability-based methods (calculating entropies or model-assigned probabilities)\cite{marek2021oodgan, xiao2021hallucination, kim2022uncertainty, stengel2023calibrated} and distance-based methods (using metrics such as the Mahalanobis distance to compare new inputs with training data)\cite{lee2018simple, zheng2020out, ren2021simple, ren2022out}. Probability-based methods gauge model confidence, enhancing output reliability. Distance-based methods identify outliers likely to produce low-quality outputs, promoting safer AI deployment. Our work employs both approaches to develop abstention mechanisms for text-to-SQL models.


\vspace{-3mm}
\section{Conclusion and Future Direction}
\vspace{-3mm}
In this work, we present \textit{TrustSQL}, a new benchmark designed to assess \textit{text-to-SQL reliability}, defined as models' ability to correctly handle any type of input question by generating correct SQL queries for feasible questions and abstaining from infeasible ones. To demonstrate this, we constructed the benchmark based on common error cases that users unfamiliar with the model's capabilities may encounter.
We developed and assessed two approaches: pipeline-based and unified methods, using existing modeling strategies. Our experimental results, using a penalty-based scoring metric, show that addressing this task requires various modeling approaches and opens new avenues for model development.

Though we have carefully designed the benchmark, there are several limitations. First, despite our efforts to cover a wide range of infeasible questions, our benchmark does not encompass all those found in real-world scenarios. Secondly, our work focuses on single-turn text-to-SQL tasks and does not address the complexities of disambiguating uncertainty in entity and schema linking or user intents. Finally, results of commercial API-based models in our experiment may not be reproducible if any updates are made.

Future research directions include developing agent-based methods that can plan ahead to achieve high performance in RS(N), bypassing the significant need for strategies such as prompt engineering and domain adaptation, accelerating wider adoption of text-to-SQL across diverse fields.



{
\small
\bibliographystyle{plain}
\bibliography{references}
}

\clearpage
\appendix
\clearpage
\startcontents[supplementary]
\printcontents[supplementary]{l}{1}{\section*{Supplementary Contents}}

\clearpage
\section{Datasheet for Datasets}
\stopcontents[supplementary]
\subsection{Motivation}

\begin{itemize}
  \item \textbf{For what purpose was the dataset created?}
  \\ TrustSQL is designed to evaluate text-to-SQL reliability—defined as a model's ability to correctly handle any type of input question by generating correct SQL queries for feasible questions and abstaining from generating infeasible ones (e.g., due to schema incompatibility or functionalities beyond SQL). By incorporating both aspects of feasible and infeasible questions in the benchmark, we aim to assess existing models' ability to gain trust in their outputs through appropriate abstention, encouraging wider adoption of text-to-SQL models across various fields beyond experimental settings.
  \item \textbf{Who created the dataset (e.g., which team, research group) and on
behalf of which entity (e.g., company, institution, organization)?}
  \\ The authors of this paper.
  \item \textbf{Who funded the creation of the dataset? If there is an associated grant, please provide the name of the grantor and the grant name and number.}
  \\ This work was supported by Institute for Information \& communications Technology Promotion(IITP) grant (No.RS-2019-II190075) and the National Research Foundation of Korea (NRF) grant (NRF-2020H1D3A2A03100945) funded by the Korea government (MSIT).
\end{itemize}

\subsection{Composition}

\begin{itemize}
  \item  \textbf{What do the instances that comprise the dataset represent (e.g., documents, photos, people, countries)?}
  \\ TrustSQL contains natural language questions and their corresponding SQL queries (if feasible); infeasible questions do not have the corresponding SQL.
  \item \textbf{How many instances are there in total (of each type, if appropriate)?}
  \\ There are about 21.4K data instances (18.9K feasible; 2.5K infeasible). 
  \item \textbf{Does the dataset contain all possible instances or is it a sample (not necessarily random) of instances from a larger set?}
  \\ This is the complete data for TrustSQL.
  \item  \textbf{What data does each instance consist of?}
  \\ TrustSQL contains two sets of data: question-SQL pairs for feasible questions and questions only for infeasible questions.
  \item  \textbf{Is there a label or target associated with each instance?}
  \\ Labels are SQL queries.
  \item  \textbf{Is any information missing from individual instances? If so, please provide a description, explaining why this information is missing (e.g., because it was unavailable). This does not include intentionally removed information, but might include, e.g., redacted text.}
  \\ N/A.
  \item  \textbf{Are relationships between individual instances made explicit (e.g., users’ movie ratings, social network links)?}
  \\ N/A.
  \item  \textbf{Are there recommended data splits (e.g., training, development/validation, testing)?}
  \\ We provide pre-split datasets.
  \item  \textbf{Are there any errors, sources of noise, or redundancies in the dataset?}
  \\ We did our best to ensure each data instance best serves its purpose without errors or noise. Any unexpected errors or noise may occur in the future, and we will be willing to correct them if detected.
  \item  \textbf{Is the dataset self-contained, or does it link to or otherwise rely on external resources (e.g., websites, tweets, other datasets)?}
  \\ The dataset is self-contained if users comply with the licenses in the downloadable link.
  \item  \textbf{Does the dataset contain data that might be considered confidential (e.g., data that is protected by legal privilege or by doctor– patient confidentiality, data that includes the content of individuals’ non-public communications)?}
  \\ N/A.
  \item  \textbf{Does the dataset contain data that, if viewed directly, might be offensive, insulting, threatening, or might otherwise cause anxiety?}
  \\ N/A.
  \item \textbf{Does the dataset relate to people?}
  \\ The EHRSQL portion is related to people.
  \item  \textbf{Does the dataset identify any subpopulations (e.g., by age, gender)?} \\
  No.
  \item  \textbf{Is it possible to identify individuals (i.e., one or more natural persons), either directly or indirectly (i.e., in combination with other data) from the dataset?}
  \\ No.
  \item \textbf{Does the dataset contain data that might be considered sensitive in any way (e.g., data that reveals race or ethnic origins, sexual orientations, religious beliefs, political opinions or union memberships, or locations; financial or health data; biometric or genetic data; forms of government identification, such as social security numbers; criminal history)?}
  \\ The EHRSQL portion of question-SQL pairs is linked to the MIMIC-IV Demo, which is already de-identified following HIPAA standards and completely openly available to anyone who complies with its license.
\end{itemize}

\subsection{Collection Process}

\begin{itemize}
  \item  \textbf{How was the data associated with each instance acquired?}
  \\ We curated and modified question-SQL pairs in TrustSQL based on the existing text-to-SQL datasets and their databases (feasible questions). Infeasible questions were manually annotated by the authors based on the existing databases.
  \item  \textbf{What mechanisms or procedures were used to collect the data (e.g., hardware apparatuses or sensors, manual human curation, software programs, software APIs)?}
  \\ We mainly used Excel, Google Sheets, and Python to collect, process, and annotate the data. In addition, we used OpenAI’s GPT-4 to generate paraphrases for each question template.
  \item  \textbf{If the dataset is a sample from a larger set, what was the sampling strategy (e.g., deterministic, probabilistic with specific sampling probabilities)?}
  \\ N/A.
  \item   \textbf{Who was involved in the data collection process (e.g., students, crowdworkers, contractors) and how were they compensated (e.g., how much were crowdworkers paid)?}
  \\ There were three parts that required human involvement in the data collection process: preprocessing question-SQL data, question paraphrasing, and infeasible question annotation for each database. In all cases, the authors conducted all the processes with the help of GPT-4 if necessary. If GPT-4 was used, the authors manually reviewed all the data to maintain high quality.
  \item  \textbf{Over what timeframe was the data collected?}
  \\ N/A.
  \item  \textbf{Were any ethical review processes conducted (e.g., by an institutional review board)?}
  \\ N/A.
  \item \textbf{Does the dataset relate to people?}
  \\ The EHRSQL portion is related to people.
  \item  \textbf{Did you collect the data from the individuals in question directly, or obtain it via third parties or other sources (e.g., websites)?}
  \\ N/A.
  \item  \textbf{Were the individuals in question notified about the data collection?}
  \\ N/A.
  \item  \textbf{Did the individuals in question consent to the collection and use of their data?}
  \\ N/A.
  \item  \textbf{If consent was obtained, were the consenting individuals provided with a mechanism to revoke their consent in the future or for certain uses?}
  \\ N/A.
  \item  \textbf{Has an analysis of the potential impact of the dataset and its use on data subjects (e.g., a data protection impact analysis) been conducted?}
  \\ The EHRSQL database we used is already de-identified.
\end{itemize}

\subsection{Preprocessing/cleaning/labeling}

\begin{itemize}
  \item  \textbf{Was any preprocessing/cleaning/labeling of the data done (e.g., discretization or bucketing, tokenization, part-of-speech tagging, SIFT feature extraction, removal of instances, processing of missing values)?}
  \\ Extensive preprocessing was done to construct TrustSQL. Details are provided in Section~\ref{appendix:data_preprocessing}. 
  \item  \textbf{Was the “raw” data saved in addition to the preprocessed/cleaned/labeled data (e.g., to support unanticipated future uses)?}
  \\ N/A.
  \item  \textbf{Is the software that was used to preprocess/clean/label the data available?}
  \\ Preprocessing, cleaning, and annotation are done using Excel, Google Sheets, and Python.
\end{itemize}

\subsection{Uses}

\begin{itemize}
  \item  \textbf{Has the dataset been used for any tasks already?}
  \\ Raw datasets are used for text-to-SQL modeling.
  \item  \textbf{Is there a repository that links to any or all papers or systems that use the dataset?}
  \\ The raw version of ATIS and Advising is available at \url{https://github.com/jkkummerfeld/text2sql-data}. EHRSQL is available at \url{https://github.com/glee4810/ehrsql-2024}. Spider is available at \url{https://github.com/taoyds/spider}.
  \item  \textbf{What (other) tasks could the dataset be used for?}
  \\ N/A.
  \item  \textbf{Is there anything about the composition of the dataset or the way it was collected and preprocessed/cleaned/labeled that might impact future uses?}
  \\ N/A.
  \item  \textbf{Are there tasks for which the dataset should not be used?}
  \\ N/A.
\end{itemize}

\subsection{Distribution}

\begin{itemize}
  \item  \textbf{Will the dataset be distributed to third parties outside of the entity (e.g., company, institution, organization) on behalf of which the dataset was created?}
  \\ No.
  \item  \textbf{How will the dataset will be distributed (e.g., tarball on website, API, GitHub)?}
  \\ The dataset is released at~\url{https://github.com/glee4810/TrustSQL}.
  \item  \textbf{When will the dataset be distributed?}
  \\ Now.
  \item  \textbf{Will the dataset be distributed under a copyright or other intellectual property (IP) license, and/or under applicable terms of use (ToU)?}
  \\ The dataset is released under the CC-BY-4.0 license.
  \item \textbf{Have any third parties imposed IP-based or other restrictions on the data associated with the instances?}
  \\ No.
  \item \textbf{Do any export controls or other regulatory restrictions apply to the dataset or to individual instances?}
  \\ No.
\end{itemize}

\subsection{Maintenance}

\begin{itemize}
  \item  \textbf{Who will be supporting/hosting/maintaining the dataset?}
  \\ The authors of the paper.
  \item  \textbf{How can the owner/curator/manager of the dataset be contacted (e.g., email address)?}
  \\ Contact the first author (\url{gyubok.lee@kaist.ac.kr}) or other authors.
  \item  \textbf{Is there an erratum?}
  \\ No.
  \item  \textbf{Will the dataset be updated (e.g., to correct labeling errors, add new instances, delete instances)?}
  \\ If any correction is needed, we plan to upload a new version.
  \item \textbf{If the dataset relates to people, are there applicable limits on the retention of the data associated with the instances (e.g., were the individuals in question told that their data would be retained for a fixed period of time and then deleted)? }
  \\ N/A
  \item \textbf{Will older versions of the dataset continue to be supported/hosted/maintained?}
  \\ We plan to maintain only the newest version.
  \item \textbf{If others want to extend/augment/build on/contribute to the dataset, is there a mechanism for them to do so?}
  \\ Contact the authors of the paper.
\end{itemize}

\resumecontents[supplementary]

\clearpage
\section{Text-to-SQL Dataset Complexity}
\label{appendix:data_complexity}

\subsection{Dataset Selection for Seed Feasible Questions}
\label{appendix:data_complexity_data_selection}


\begin{table*}[h!]
\centering
\renewcommand{\arraystretch}{1.5}
\begin{adjustbox}{width=\textwidth,center}
\begin{tabular}{cccccccc}
\hline
Type & Dataset & \# Table/DB & \# Column/DB & \# Sample & Avg. SQL Length & Included in TrustSQL \\
\hline
\multirow{11}{*}{\makecell{Single-\\domain}} & ATIS & 25 & 131 & 5,280 & 129.7 & \checkmark \\
& GeoQuery & 7 & 29 & 877 & 33.3 &  \xmark \\
& Restaurants & 3 & 12 & 378 & 41.4 & \xmark \\
& Academic & 15 & 42 & 196 & 48.1 & \xmark \\
& Yelp & 7 & 38 & 128 & 38.5 & \xmark \\
& IMDB & 16 & 65 & 131 & 39.1 & \xmark \\
& Scholar & 12 & 28 & 817 & 49.9 & \xmark \\
& Advising & 12 & 84 & 4,387 & 65.2 & \checkmark \\
& MIMICSQL & 5 & 49 & 10,000 & 40.7 & \xmark \\
& SEDE & 29 & 211 & 12,023 & 74.5 & \xmark \\
& EHRSQL & 17 & 113 & 10,083 & 74.4 & \checkmark \\
\hline
\multirow{4}{*}{\makecell{Cross-\\domain}} & WikiSQL & 1.0 & 6.3 & 80,654 & 17.8 & \xmark \\
& Spider & 5.1 & 26.8 & 8,034 & 18.5 & \checkmark \\
& KaggleDBQA & 2.3 & 22.5 & 272 & 17.3 & \xmark \\
& BIRD & 7.5 & 54.2 & 10,962 & 34.4 & \xmark \\
\hline
\end{tabular}
\end{adjustbox}
\caption{Open-source text-to-SQL datasets in the literature.}
\end{table*}

\noindent 
\paragraph{Single-domain} 
We select datasets that feature complex SQL queries (Avg. SQL Length) and database schemas (\# Table/DB and \# Column/DB). SEDE~\cite{hazoom2021sede} is excluded from our dataset selection because its SQL queries are too inconsistent (raw queries written by real users at Stack Exchange) and lack accompanying executable databases. As a result, we selected ATIS, Advising, and EHRSQL, representing diverse domains with complex queries.

\paragraph{Cross-domain} We select Spider over other datasets for the following reasons: WikiSQL is limited by single-table questions (\# Table/DB) and simplified SQL queries. KaggleDBQA offers realistic questions based on actual Web databases, but its schema is considered simple (\# Table/DB) for our needs. BIRD features questions and complex SQL queries with their mapping evidence for SQL generation. However, this reliance on evidence diverges from our objective to include both feasible and infeasible questions, leading us to exclude it from our dataset selection. Therefore, Spider is the most suitable choice for our research.

\subsection{Query Difficulty in TrustSQL}
\label{appendix:data_complexity_sample_complexity}

For assigning query difficulty \textbf{[P1]}, we adopted the SQL hardness criteria following DIN-SQL~\cite{pourreza2023din-sql} across all datasets, classifying samples into three levels: \texttt{easy}, \texttt{medium}, and \texttt{hard}. The \texttt{easy} category includes single-table queries that do not require joins or nesting. The \texttt{medium} category includes queries that involve joins (including implicit joins) but excludes nesting. The \texttt{hard} category covers queries that can contain joins, sub-queries, and set operations.

Table~\ref{tab:hardness_statistic} shows the data statistics for each hardness level for feasible questions. Infeasible questions do not have hardness levels but can be categorized into five question types (statistics in Table~\ref{tab:data_statistic}) used for their annotations. Samples of feasible questions and their corresponding SQL queries by hardness level are in Table~\ref{tab:feasible_sample}.

\begin{table*}[h!]
\centering
\caption{Data statistics for feasible questions by hardness level. EHRSQL contains no medium-level samples because its annotation process avoided using joins and used nesting instead.}
\renewcommand{\arraystretch}{1.5}
\begin{adjustbox}{width=\textwidth,center}
\begin{tabular}{ccccccccccccccccc}
\hline

\multirow{2.0}{*}{split} & \multicolumn{4}{c}{ATIS} & \multicolumn{4}{c}{Advising} & \multicolumn{4}{c}{EHRSQL} 
& \multicolumn{4}{c}{Spider} \\

\cmidrule(lr){2-5}
\cmidrule(lr){6-9}
\cmidrule(lr){10-13}
\cmidrule(lr){14-17}

& total & easy & medium & hard
& total & easy & medium & hard
& total & easy & medium & hard
& total & easy & medium & hard \\
\hline

total
& 2079 & 159 & 1361 & 559
& 2236 & 328 & 1578 & 330
& 6539 & 688 & - & 5851 
& 8034 & 4115 & 2741 & 1178 \\
\hline

train
& 1114 & 83 & 735 & 296
& 1170 & 158 & 846 & 166
& 4674 & 452 & - & 4222 
& 7000 & 3571 & 2410 & 1019 \\

valid
& 489 & 37 & 320 & 132
& 533 & 84 & 368 & 81
& 931 & 114 & - & 817 
& 507 & 264 & 175 & 68 \\

test
& 476 & 39 & 306 & 131
& 533 & 86 & 364 & 83
& 934 & 122 & - & 812 
& 527 & 280 & 156 & 91 \\
\hline
\end{tabular}
\end{adjustbox}
\label{tab:hardness_statistic}
\end{table*}

\begin{table}[t!]
\caption{Sample feasible questions. SQL queries in domain-specific datasets are very long, posing a significant challenge in handling lengthy queries for the model. In contrast, those in cross-domain datasets are relatively short but require database generalization. EHRSQL does not have any medium-difficulty samples.}
\centering
\renewcommand{\arraystretch}{2.0}
\begin{adjustbox}{width=\columnwidth,center}  
\begin{tabular}{ccll}
\toprule
\textbf{Database} & \textbf{Difficulty} & \multicolumn{1}{c}{\textbf{Question}} & \multicolumn{1}{c}{\textbf{SQL Query}} \\

\midrule
\multirow{5}{*}{ATIS} 
& easy
& Can you inform me about the location of LGA?
& \makecell[l]{SELECT DISTINCT AIRPORTalias0.AIRPORT\_LOCATION FROM \\
AIRPORT AS AIRPORTalias0 WHERE \\
AIRPORTalias0.AIRPORT\_CODE = "LGA"} \\

\cmidrule(lr){2-4}
& medium
& Can you provide a list of flights that arrive at DAL?
& \makecell[l]{SELECT DISTINCT FLIGHTalias0.FLIGHT\_ID FROM AIRPORT AS \\
AIRPORTalias0 , FLIGHT AS FLIGHTalias0 WHERE \\
AIRPORTalias0.AIRPORT\_CODE = "DAL" AND \\
FLIGHTalias0.TO\_AIRPORT = AIRPORTalias0.AIRPORT\_CODE} \\

\cmidrule(lr){2-4}
& hard
& \makecell[l]{What would be the cheapest flight from ATLANTA \\
to DENVER on 10 / 12 / 1991?}
& \makecell[l]{SELECT DISTINCT
SELECT DISTINCT FLIGHTalias0.FLIGHT\_ID \\
FROM AIRPORT\_SERVICE AS AIRPORT\_SERVICEalias0 , \\ 
\textbf{... [38 lines omitted involving query nesting] ...} \\
FLIGHTalias0.FLIGHT\_ID = FLIGHT\_FAREalias0.FLIGHT\_ID} \\


\midrule
\multirow{6.5}{*}{Advising} 
& easy
& \makecell[l]{Give me the course number of the Investigations \\ class.}
& \makecell[l]{SELECT DISTINCT COURSEalias0.NUMBER FROM COURSE AS \\
COURSEalias0 WHERE COURSEalias0.NAME LIKE \\ "\%Investigations\%"} \\

\cmidrule(lr){2-4}
& medium
& \makecell[l]{Please provide me with the PreMajor courses that \\ were available in Fall 2015.}
& \makecell[l]{SELECT DISTINCT COURSEalias0.DEPARTMENT , \\
COURSEalias0.NUMBER FROM COURSE AS COURSEalias0 \\
INNER JOIN COURSE\_OFFERING AS COURSE\_OFFERINGalias0 \\
\textbf{... [9 lines omitted involving joins] ...} \\
= 2015} \\

\cmidrule(lr){2-4}
& hard
& \makecell[l]{Who, most recently, taught SCAND 104 before \\ 2016?}
& \makecell[l]{SELECT DISTINCT INSTRUCTORalias0.NAME FROM COURSE AS \\
COURSEalias0 INNER JOIN COURSE\_OFFERING AS \\
COURSE\_OFFERINGalias0 ON COURSEalias0.COURSE\_ID = \\
\textbf{... [18 lines omitted involving query nesting] ...} \\
COURSEalias0.NUMBER = 104} \\

\midrule
\multirow{5}{*}{EHRSQL}
& easy
& \makecell[l]{What are the ways to consume metformin \\ (glucophage)?}
& \makecell[l]{SELECT DISTINCT prescriptions.route FROM prescriptions WHERE \\ 
prescriptions.drug = 'metformin (glucophage)'} \\
\cmidrule(lr){2-4}

& medium 
& \makecell[l]{-}
& \makecell[l]{-} \\
\cmidrule(lr){2-4}

& hard & 
\makecell[l]{What was the change in arterial blood pressure \\ systolic in patient 10037975 second measured on \\ the first ICU visit compared to the first value \\ measured on the first ICU visit?}
& \makecell[l]{SELECT ( SELECT chartevents.valuenum FROM chartevents \\ 
WHERE chartevents.stay\_id IN ( SELECT icustays.stay\_id FROM \\
icustays WHERE icustays.hadm\_id IN ( SELECT admissions.hadm\_id \\
\textbf{... [15 lines omitted involving query nesting] ...} \\
ASC LIMIT 1 )} \\

\midrule
\multirow{4.5}{*}{SPIDER} 
& easy
& What is the first and second line for all addresses?
& SELECT line\_1 ,  line\_2 FROM addresses \\

\cmidrule(lr){2-4}
& medium 
& \makecell[l]{What is the average earnings of poker players with \\ height higher than 200?}
& \makecell[l]{SELECT avg(T2.Earnings) FROM people AS T1 JOIN poker\_player \\ AS T2 ON T1.People\_ID = T2.People\_ID WHERE T1.Height  >  200} \\

\cmidrule(lr){2-4}
& hard & 
\makecell[l]{Find the first name of student who is taking classes \\ from accounting and Computer Info. Systems \\ departments.}
& \makecell[l]{SELECT T1.stu\_fname FROM student AS T1 JOIN enroll AS T2 ON \\
T1.stu\_num  =  T2.stu\_num JOIN CLASS AS T3 ON T2.class\_code  = \\
\textbf{... [7 lines omitted involving set operations] ...} \\
=  'Computer Info. Systems'} \\

\bottomrule
\end{tabular}
\end{adjustbox}
\label{tab:feasible_sample}
\end{table}

\clearpage
\section{Modeling Details of Open-Source Models}
\label{appendix:modeling_detail_opensource}

\subsection{Pipeline-based Approach}

\paragraph{SQL Generation (\textsc{T5})} We use \texttt{T5-3B} as the backbone model. The input consists of questions followed by a serialized schema listing all tables and columns for each database~\cite{suhr-etal-2020-exploring}. The model is trained using BF16 with the Adam optimizer at a learning rate of 0.0001, continuing until the validation loss stops decreasing. Each model is independently trained on four different datasets using NVIDIA RTX A6000 GPUs.
\paragraph{Infeasible Question Detection (\textsc{CLS$_M$})} We fine-tune \texttt{sqlcoder-7b-2} for classifying question types (small talk, ambiguous, lack of data, unanswerable by SQL, and answerable) using the TriageSQL~\cite{triagesql} training set. This task is trained for one epoch on NVIDIA RTX A6000 GPUs and evaluated on TrustSQL's test set, categorizing answerable questions as feasible and others as infeasible.
\paragraph{SQL Error Detection (\textsc{Error$_M$})} We also utilize \texttt{sqlcoder-7b-2} for detecting SQL errors. The training data is derived from the sample-level accuracy of SQL generation for TrustSQL's validation set, combined with sampled infeasible data from TriageSQL (the same number of data points as the TrustSQL validation set). Correct labels correspond to accurate SQL generation for feasible questions in the TrustSQL validation set, while incorrect labels include both inaccurate SQL generation for feasible questions in the TrustSQL validation set and infeasible questions from TriageSQL. This task is also trained for one epoch on NVIDIA RTX A6000 GPUs.

\subsection{Unified Approach}
\label{appendix:thresholding}

We leverage the same \textsc{T5} model used in the pipeline-based approach for the unified approach. The methods for abstention in this approach involve probability-based (\textit{i.e.}, \textsc{MaxEnt} and \textsc{MaxProb}) and distance-based (\textit{i.e.}, \textsc{FeatMD} and \textsc{FeatRMD}) methods, which require fixed thresholds to draw the boundary between SQL generation and abstention from the model outputs.

\begin{figure}[h!]
\centering
\includegraphics[width=\linewidth]{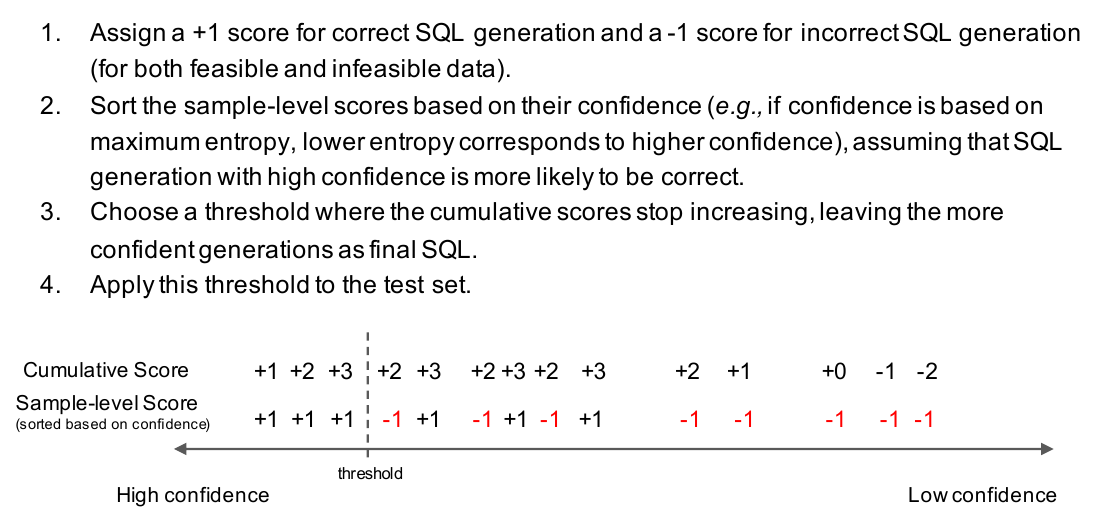}
\caption{Samples sorted in descending confidence. 14 samples are illustrated in the figure.}
\label{fig:thresholding}
\end{figure}

\begin{figure*}
    \centering
    \begin{subfigure}[b]{0.475\textwidth}
        \centering
        \includegraphics[width=\textwidth]{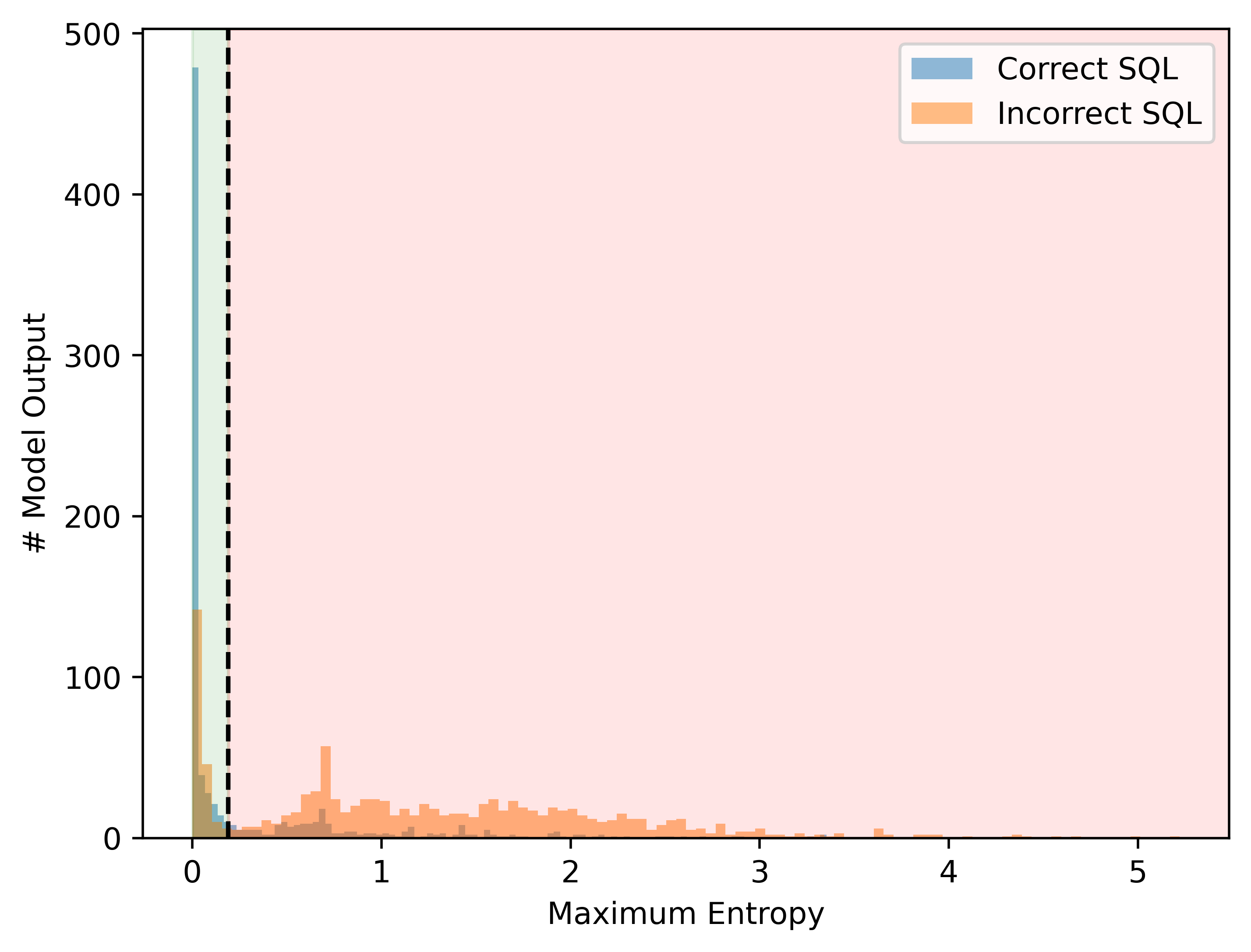}
        \caption{\textsc{T5[MaxEnt]}} 
        \label{fig:mean and std of net14}
    \end{subfigure}
    \hfill
    \begin{subfigure}[b]{0.475\textwidth}  
        \centering 
        \includegraphics[width=\textwidth]{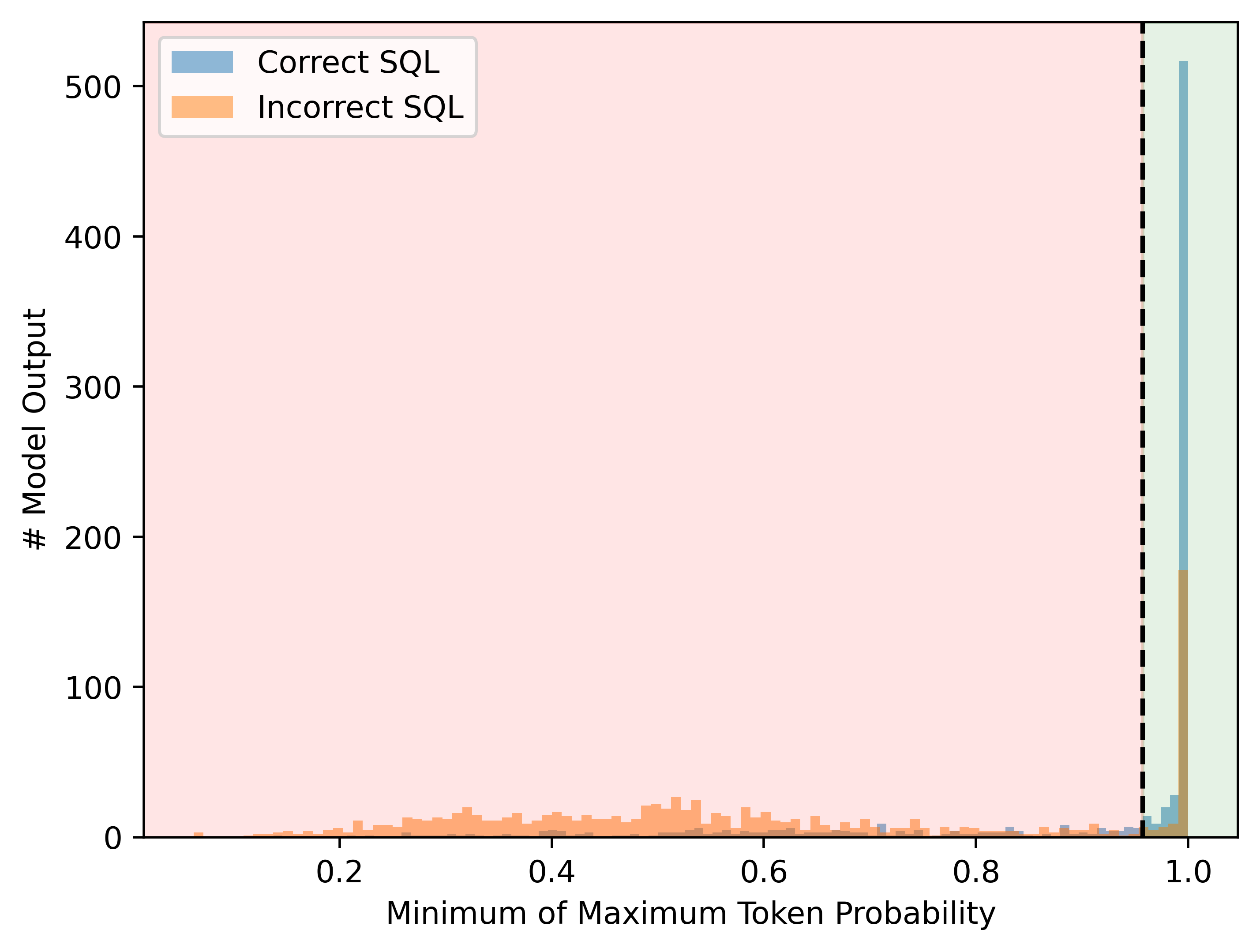}
        \caption{\textsc{T5[MaxProb]}}    
        \label{fig:mean and std of net24}
    \end{subfigure}
    \vskip\baselineskip
    \begin{subfigure}[b]{0.475\textwidth}   
        \centering 
        \includegraphics[width=\textwidth]{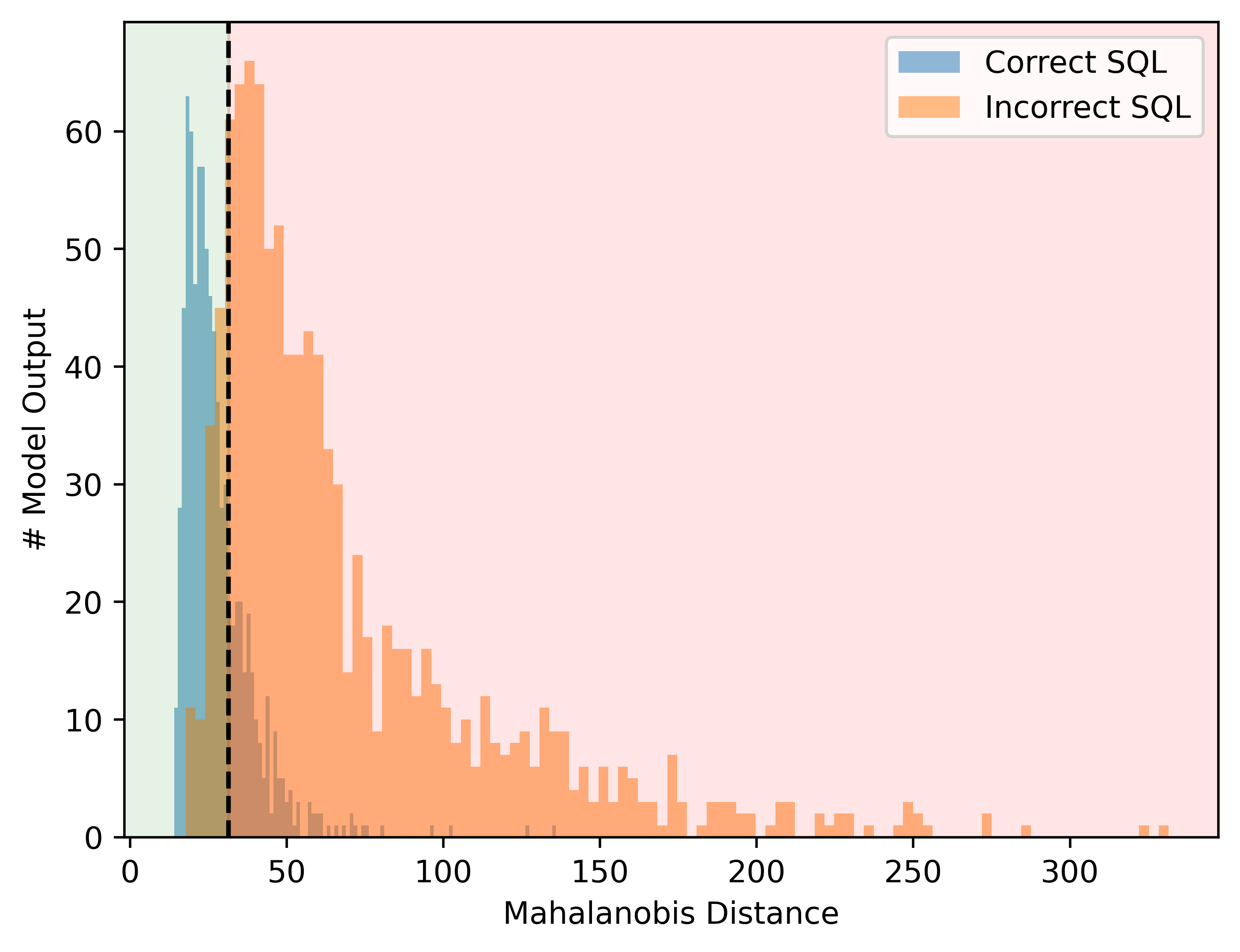}
        \caption{\textsc{T5[FeatMD]}}
        \label{fig:mean and std of net34}
    \end{subfigure}
    \hfill
    \begin{subfigure}[b]{0.475\textwidth}   
        \centering 
        \includegraphics[width=\textwidth]{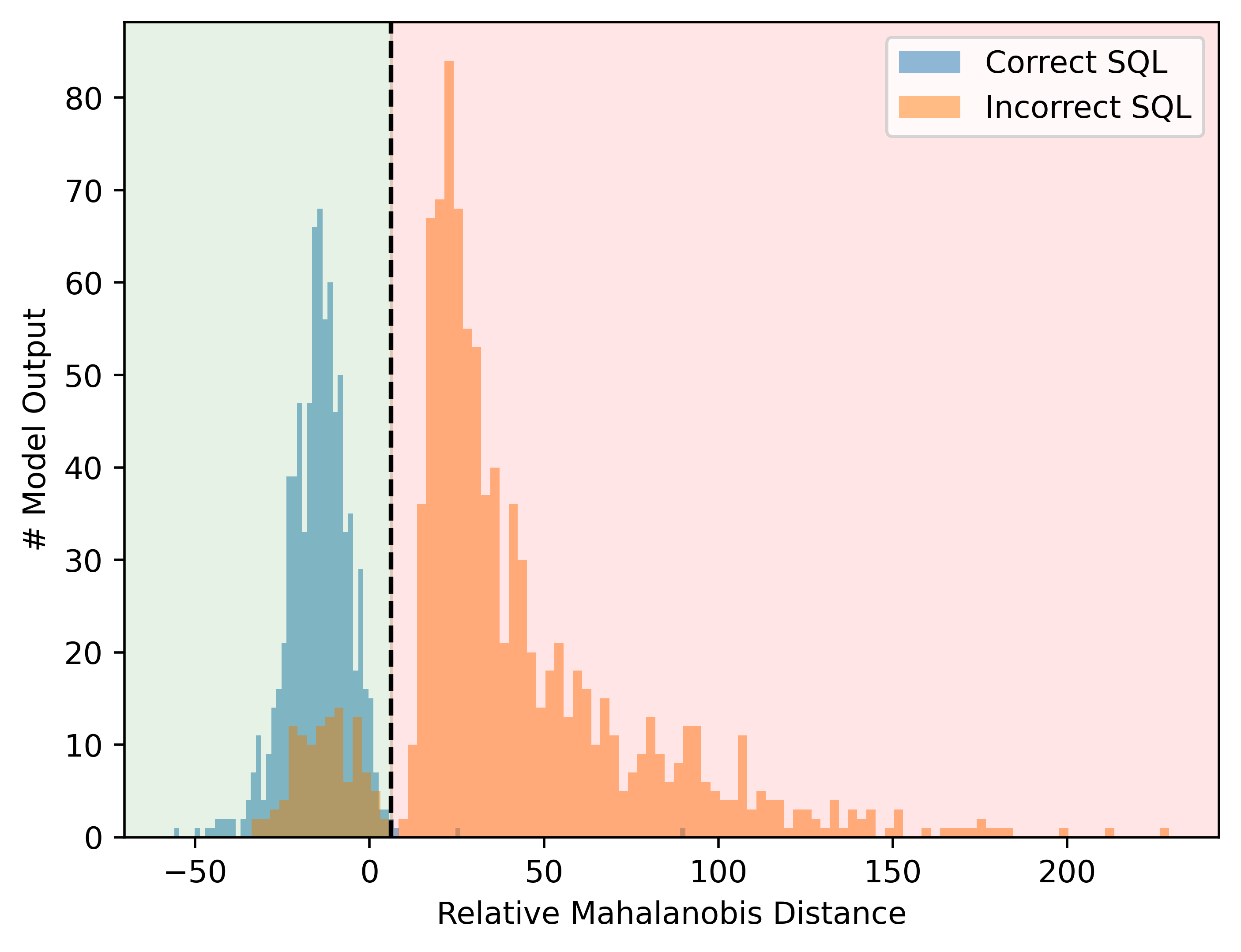}
        \caption{\textsc{T5[FeatRMD]}}
        \label{fig:mean and std of net44}
    \end{subfigure}
    \caption{Visualization of thresholds chosen for the EHRSQL portion in the validation set. The green region in each plot is considered confident by the model, so samples that fall under this region are used for final SQL generation, while the red region is not confident and thus the samples under this region are abstained from.}
    \label{fig:sample_visualized_threshold}
\end{figure*}

\paragraph{Threshold Selection Method} Figure~\ref{fig:thresholding} shows the steps for our heuristic threshold method, and Figure~\ref{fig:sample_visualized_threshold} presents the visualization of the thresholds for \textsc{MaxEnt}, \textsc{MaxProb}, \textsc{FeatMD}, and \textsc{FeatRMD}.

\clearpage
\section{Modeling Details of Commercial API-based Models}
\label{appendix:modeling_detail_api}

\subsection{Pipeline-based Approach}

\paragraph{SQL Generation}

For all methods described below, we used GPT-3.5 (\texttt{gpt-3.5-turbo-0125}) instead of GPT-4 due to budget constraints. We considered using GPT-4o, but the proposed prompts for \textsc{DIN-SQL} and \textsc{MAC-SQL} were not compatible with the new model in a stable manner, particularly due to model output formatting issues. Therefore GPT-4o is used only for \textsc{CLS$_{P}$} and \textsc{Error$_{P}$}.

\begin{itemize}
    \item \textsc{SQLPrompt}~\cite{chang2023prompt} We use the same schema representation (\texttt{Columns=[]+FK}) as \textsc{DIN-SQL}. For question representation following the schema, we use a retrieval-based prompt (8 samples x 1 database) with a dense retriever (\texttt{all-mpnet-base-v2}) for the single-domain settings, while a fixed prompt (8 samples x 5 databases, randomly selected) is used for the cross-database setting.
     \item \textsc{DIN-SQL}~\cite{pourreza2023din-sql} We adjust prompts for single-domain datasets, except for the self-correction module at the end. This step is necessary because the original prompts are limited to the SQL structures and syntax found in Spider, making them insufficient for handling the complexity of domain-specific text-to-SQL data.
    \item \textsc{MAC-SQL}~\cite{wang2023mac} We adjust the selector and decomposer prompts, except for the refiner, for both single-domain and cross-domain datasets according to their database schemas and samples as task demonstrations.
\end{itemize}

\paragraph{Infeasible Question Detection (\textsc{CLS$_{P}$})}

We use \texttt{CreateTable + SelectCol3} to represent database schemas. This method has shown the best performance among various schema representation choices~\citep{chang2023prompt}. For the single-domain settings, we utilize 16 feasible questions (retrieved from both the TrustSQL training and validation sets) and 16 infeasible questions (retrieved from TriageSQL) with a dense retriever (\texttt{all-mpnet-base-v2}), following the schema representation. For the cross-domain setting, we utilize two feasible and two infeasible questions (retrieved from TriageSQL) from 8 different random databases.



\paragraph{SQL Error Detection (\textsc{Error$_{p}$})} This model is designed to hold back the model output if any errors exist in the generated SQL. The prompt is a modified version of the self-correction module in DIN-SQL~\cite{pourreza2023din-sql}, starting from \texttt{Columns=[]+FK} as schema representation, followed by the input question and generated SQL. The instruction is: \textit{Based on the question and predicted SQL, are you sure the SQL below is correct? If you consider the SQL to be correct, answer with `correct'. If not, answer with `incorrect'. Only output your response without explanation.}


\subsection{Unified Approach}

Based on the prompt used in \textsc{SQLPrompt}, \textsc{SQLPrompt[Demo]} uses the same number of infeasible questions as a few-shot demonstration (these samples are randomly shuffled with feasible ones), while \textsc{SQLPrompt[Voting]} uses exactly the same prompt as \textsc{SQLPrompt} but is sampled five times to check for unanimity in SQL generation; otherwise, it abstains.

\clearpage
\section{Qualitative Analysis for Reliability Scores}
\label{appendix:rs_qualitative}

\begin{figure}[h!]
\centering

\begin{subfigure}{\textwidth}
  \centering
  \includegraphics[width=0.9\linewidth]{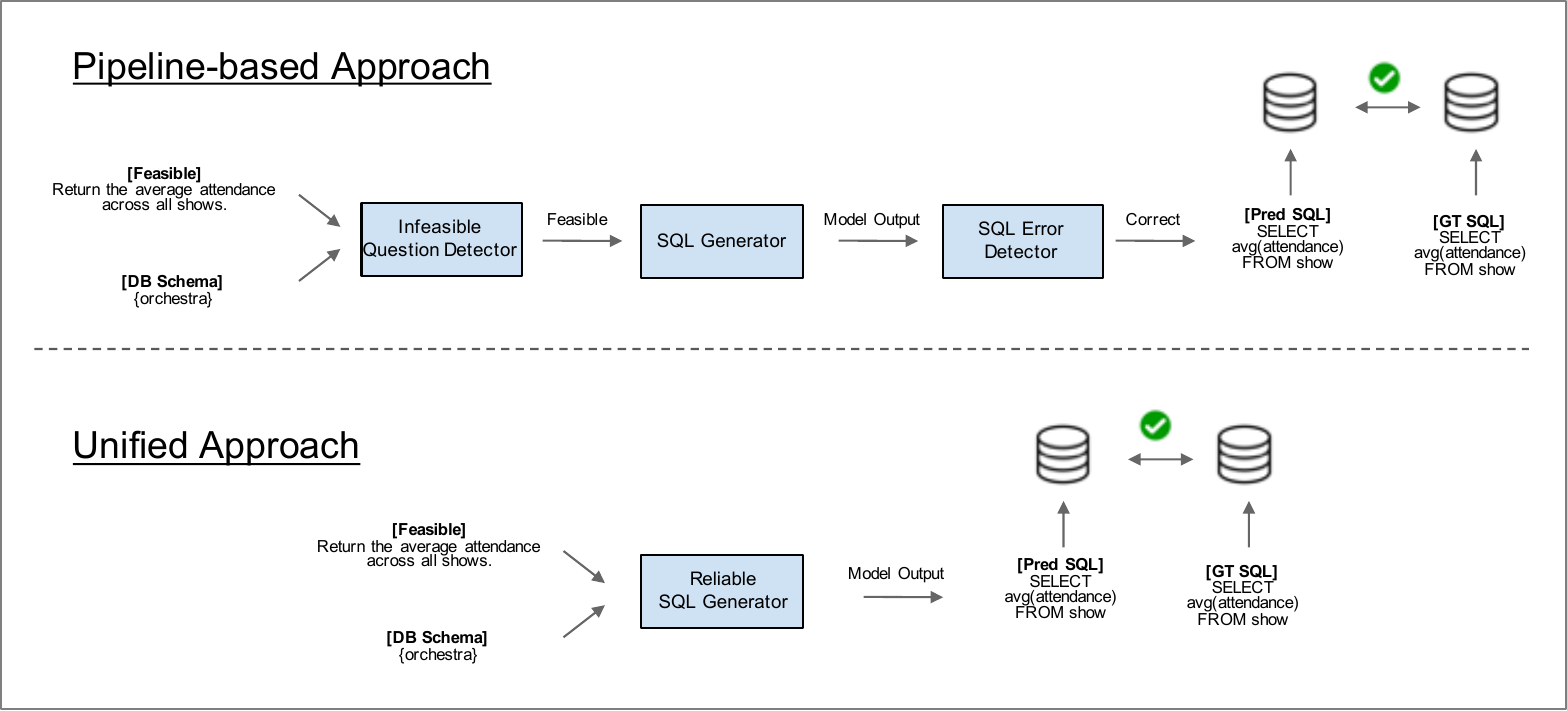}
  \caption{Example of \textbf{(I)}. A correct SQL query is generated for a feasible question.} 
\end{subfigure}\\[5ex]
\begin{subfigure}{\textwidth}
  \centering
  \includegraphics[width=0.9\linewidth]{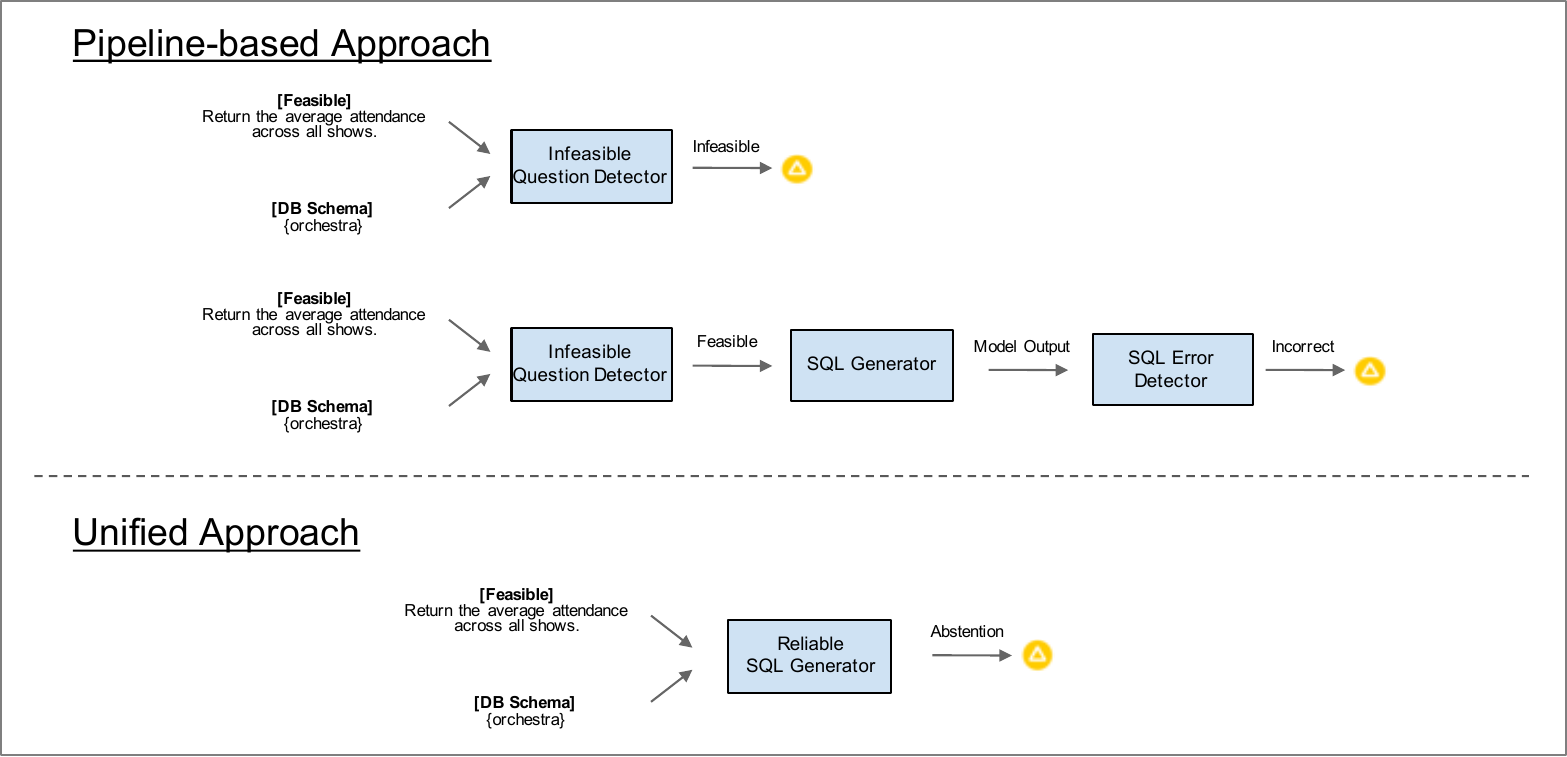}
  \caption{Example of \textbf{(II)}. SQL generation is abstained for a feasible question.}
\end{subfigure}\\[5ex]
\begin{subfigure}{\textwidth}
  \centering
  \includegraphics[width=0.9\linewidth]{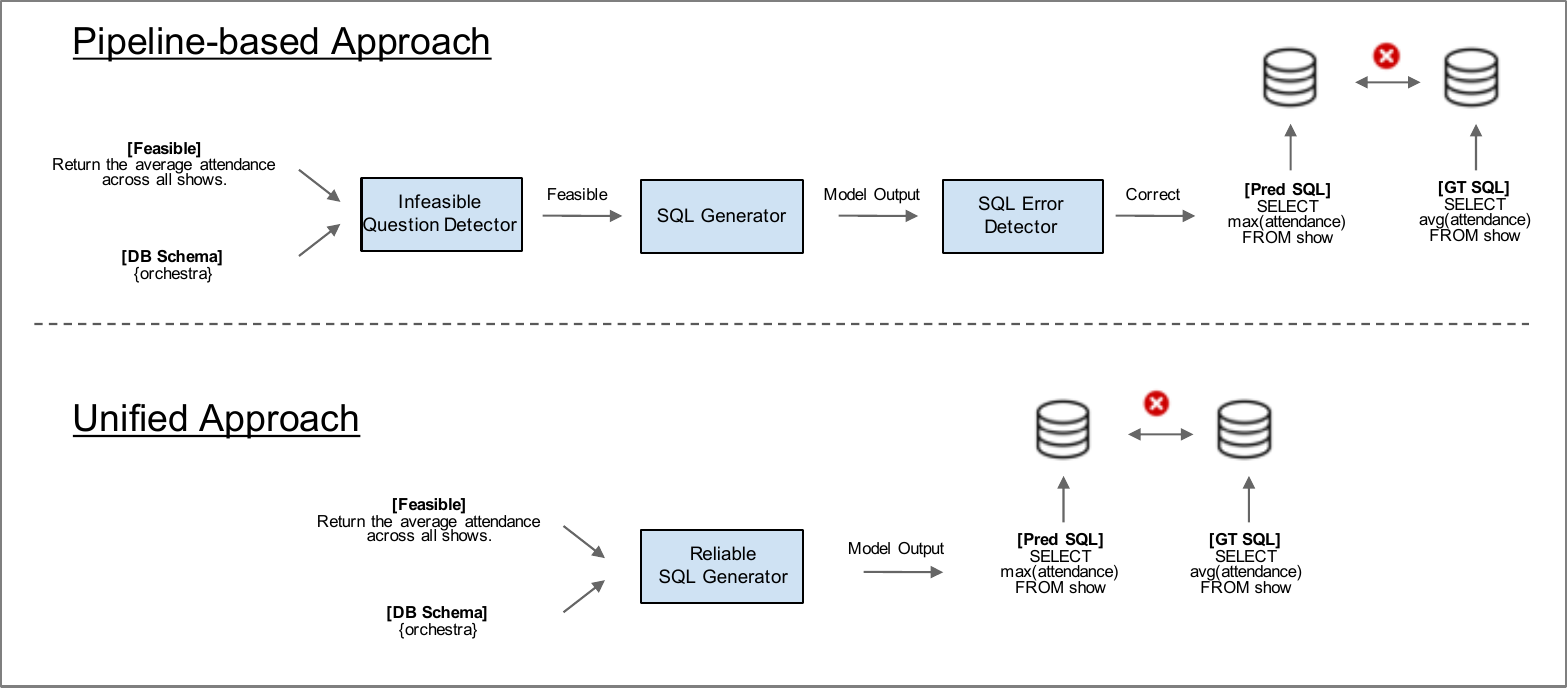}
  \caption{Example of \textbf{(III)}. An incorrect SQL is generated for a feasible question.}  
\end{subfigure}
\end{figure}

\begin{figure}[h!]
\centering

\begin{subfigure}{\textwidth}
  \centering
  \includegraphics[width=0.9\linewidth]{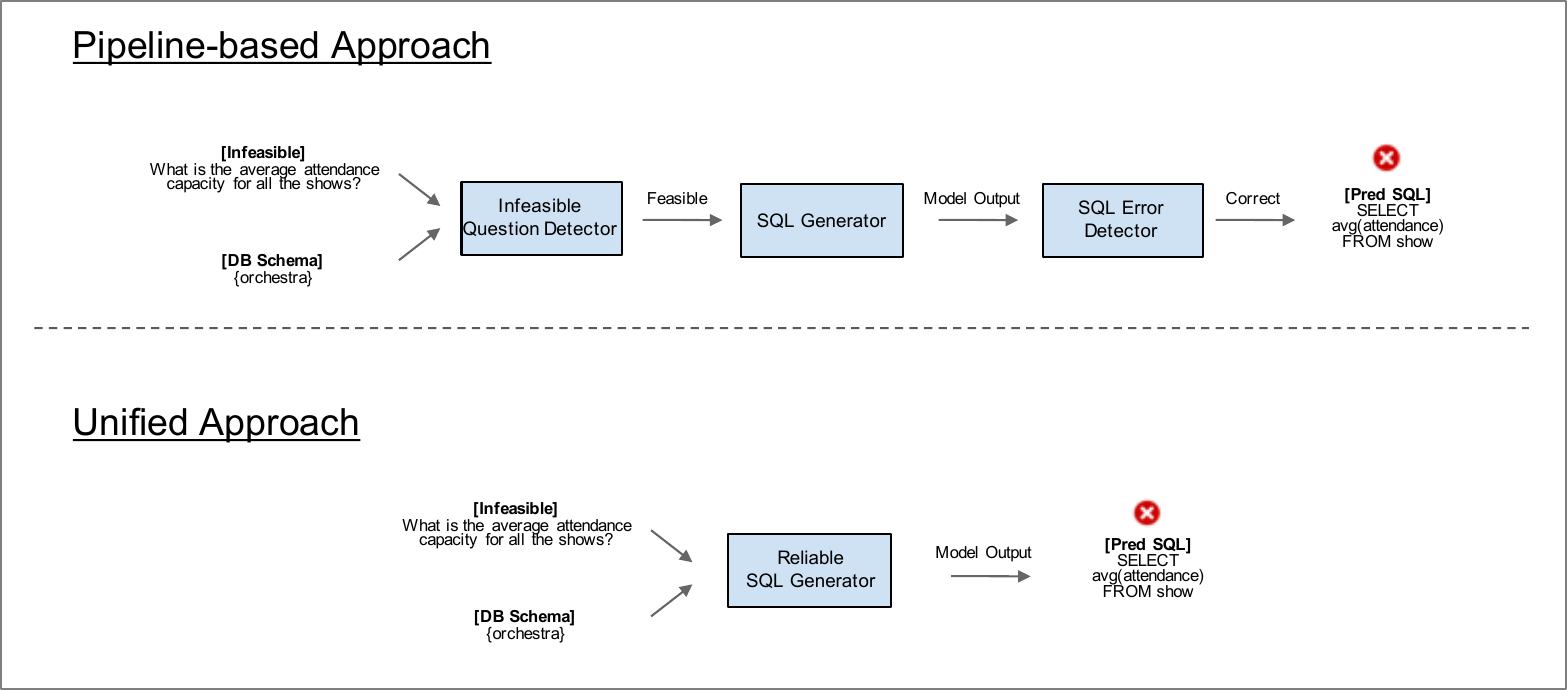}
  \caption*{(d) Example of \textbf{(IV)}. A SQL query is generated for an infeasible question.}
\end{subfigure}\\[5ex]
\begin{subfigure}{\textwidth}
  \centering
  \includegraphics[width=0.9\linewidth]{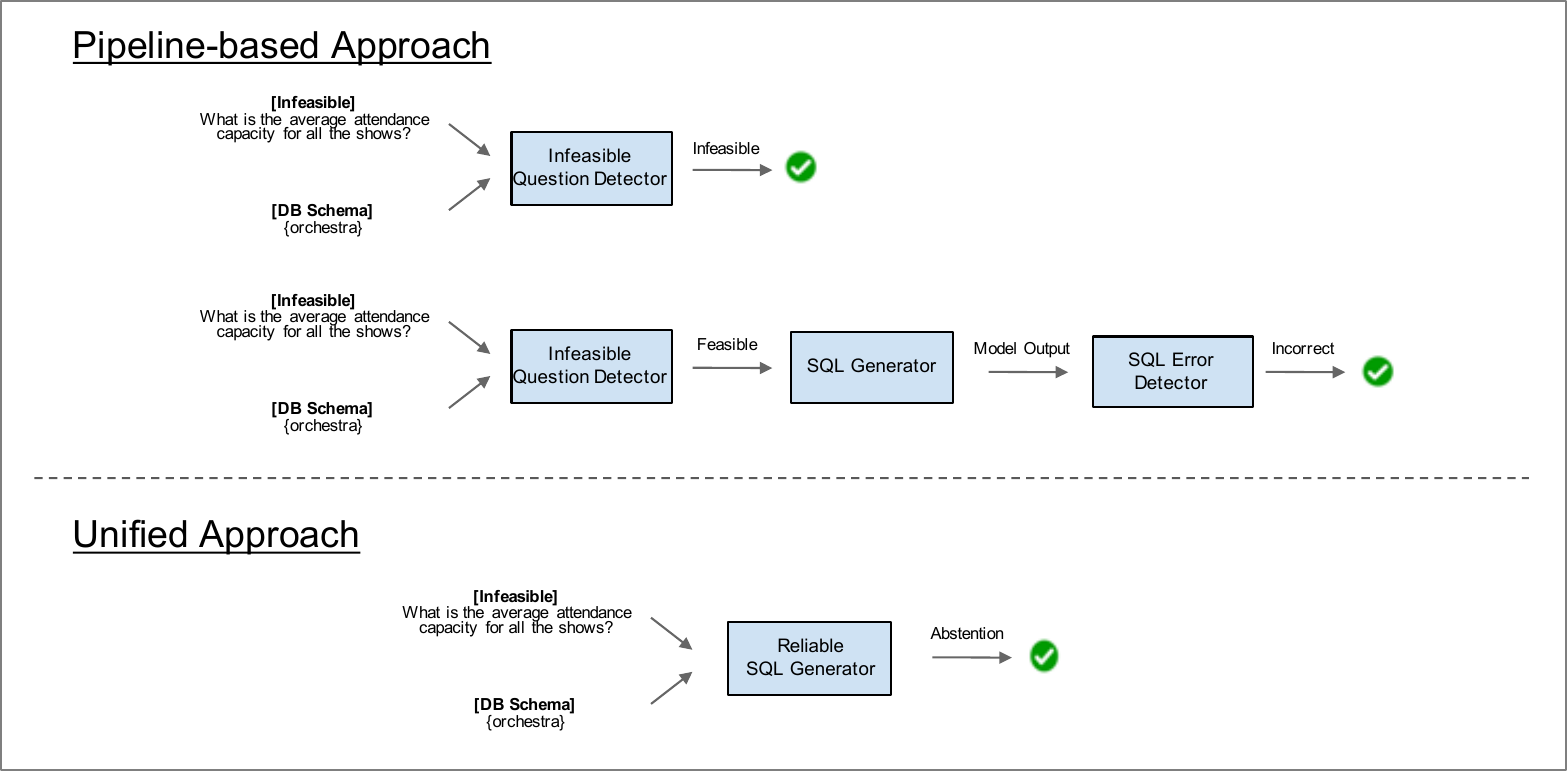}
  \caption*{(e) Example of \textbf{(V)}. An infeasible question is properly abstained.}
\end{subfigure}
\end{figure}






\clearpage
\section{Detailed Model Performance}
\label{appendix:error_analysis}

\subsection{Performance of Correct SQL among all Model Outputs}

While we evaluate model performance in terms of the green region over the whole region in Figure~\ref{fig:sub3} \textbf{(I+V)/(I+II+III+IV+V)} and through the scores provided in Table~\ref{tab:main_all} under different penalties, we take a different view: the ratio of correct SQL to all SQL generation (\textit{i.e.}, model output without abstention) \textbf{(I)/(I+III+IV)}, as shown in Table~\ref{tab:correct_sql_generation}. In high-penalty scenarios, SQL generation should be done conservatively only if it is certain, and this ratio inversely indicates the extent of incorrect SQL generation going unnoticed by each model.

\begin{table*}[h!]
\centering
\renewcommand{\arraystretch}{1.8}
\begin{adjustbox}{width=\textwidth,center}

\begin{tabular}{ccccccc}
\hline

\multirow{2.5}{*}{}
& & & \multicolumn{1}{c}{ATIS} & \multicolumn{1}{c}{Advising} & \multicolumn{1}{c}{EHRSQL} & \multicolumn{1}{c}{Spider} \\
\hline
\multirow{8}{*}{Open}
& \multirow{3}{*}{Pipeline}
& \textsc{CLS$_M$} → \textsc{T5} & \makecell{42.9 \\ 155 / (260+101)} & \makecell{41.1 \\ 151 / (196+171)} & \makecell{63.4 \\ 398 / (468+160)} & \makecell{40.1 \\ 258 / (502+141)}  \\ 
& & \textsc{T5 → Error$_M$} & \makecell{68.5 \\ 274 / (345+55)} & \makecell{63.0 \\ 76 / (475+122)} & \makecell{67.4 \\ 738 / (833+262)} & \makecell{53.2 \\ 218 / (301+109)} \\
& & \textsc{CLS$_M$ → T5 → Error$_M$} & \makecell{\textbf{76.2} \\ 154 / (184+18)} & \makecell{63.7 \\ 151 / (181+56)} & \makecell{78.6 \\ 357 / (406+48)} & \makecell{\underline{60.3} \\ 210 / (289+59)} \\
\cdashline{2-7}
& \multirow{4}{*}{Unified}
& \textsc{T5[MaxEnt]} & \makecell{73.3 \\ 151 / (186+20)} & \makecell{\textbf{85.2} \\ 281 / (311+19)} & \makecell{\underline{88.7} \\ 725 / (752+65)} & \makecell{55.2 \\ 169 / (224+82)} \\
& & \textsc{T5[MaxProb]} & \makecell{\underline{75.0} \\ 132 / (154+22)} & \makecell{\underline{82.7} \\ 286 / (320+26)} & \makecell{\textbf{88.8} \\ 725 / (751+65)} & \makecell{59.7 \\ 117 / (150+46)} \\
& & \textsc{T5[FeatMD]} & \makecell{69.8 \\ 169 / (240+2)} & \makecell{46.5 \\ 345 / (473+269)} & \makecell{79.5 \\ (387 / (429+58))} & \makecell{\textbf{100.0} \\ 1 / (1+0)} \\ 
& & \textsc{T5[FeatRMD]} & \makecell{65.5 \\ 218 / (332+1)} & \makecell{45.4 \\ 361 / (497+299)} & \makecell{65.3 \\ 808 / (923+314)} & \makecell{35.1 \\ 197 / (364+198)} \\
\hline
\hline
\multirow{8}{*}{API}
& \multirow{5}{*}{Pipeline}
& \textsc{CLS$_P$ → SQLPrompt} & \makecell{81.2 \\ 376 / (457+6)} & \makecell{\underline{81.3} \\ 427 / (511+14)} & \makecell{75.4 \\ 764 / (909+104)} & \makecell{49.5 \\ 200 / (384+20)} \\
& & \textsc{CLS$_P$ → DIN-SQL} & \makecell{24.8 \\ 115 / (457+6)} & \makecell{54.9 \\ 288 / (511+14)} & \makecell{31.6 \\ 321 / (909+107)} & \makecell{\textbf{70.8} \\ 294 / (384+31)} \\
& & \textsc{CLS$_P$ → MAC-SQL} & \makecell{49.0 \\ 227 / (457+6)} & \makecell{45.3 \\ 238 / (511+14)} & \makecell{39.9 \\ 405 / (909+105)} & \makecell{\underline{69.3} \\ 287 / (384+30)} \\
& & \textsc{SQLPrompt → Error$_P$} & \makecell{63.5 \\ 47 / (50+24)} & \makecell{65.6 \\ 103 / (116+41)} & \makecell{64.3 \\ 151 / (165+70)} & \makecell{58.6 \\ 267 / (422+34)} \\
& & \textsc{CLS$_P$ → SQLPrompt → Error$_P$} & \makecell{\textbf{93.5} \\ 43 / (46+0)} & \makecell{\textbf{87.4} \\ 97 / (109+2)} & \makecell{\underline{88.2} \\ 150 / (158+12)} & \makecell{61.9 \\ 187 / (301+1)} \\
\cdashline{2-7}
& \multirow{2}{*}{Unified}
& \textsc{SQLPrompt[Demo]} & \makecell{60.1 \\ 378 / (474+155)} & \makecell{57.8 \\ 432 / (533+215)} & \makecell{57.0 \\ 768 / (927+420)} & \makecell{52.9 \\ 393 / (527+216)} \\
& & \textsc{SQLPrompt[Voting]} & \makecell{\underline{87.2} \\ 164 / (178+10)} & \makecell{\textbf{87.4} \\ 202 / (215+16)} & \makecell{\textbf{89.7} \\ 463 / (490+26)} & \makecell{53.2 \\ 25 / (29+18)} \\
\hline
\end{tabular}
\end{adjustbox}
\caption{The first line of each cell indicates the ratio of correct SQL to all SQL generations. The second line indicates the number of correct SQL divided by the sum of generated SQL without abstention for both feasible and infeasible questions.}
\label{tab:correct_sql_generation}
\end{table*}

The results in Table~\ref{tab:correct_sql_generation} that (1) unified methods tend to have higher ratios of correct SQL per total number of generations compared to pipeline-based methods, and (2) models that are conservative (with low values in the denominators) do not necessarily have the highest ratios of correct SQL.

\clearpage
\subsection{Performance of Infeasible Question Detection for Pipeline-based Methods}

Tables~\ref{tab:infeasible_detection_feasible} and~\ref{tab:infeasible_detection_infeasible} show the performance of infeasible question detection by pipeline-based methods.

\begin{table*}[h!]
\centering
\renewcommand{\arraystretch}{1.5}
\begin{adjustbox}{width=\textwidth,center}
\begin{tabular}{ccccccccccccc}
\hline

\multirow{2.5}{*}{}
 & \multicolumn{2}{c}{ATIS} & \multicolumn{2}{c}{Advising} & \multicolumn{2}{c}{EHRSQL} & \multicolumn{1}{c}{Spider}\\

\cmidrule(lr){2-3}
\cmidrule(lr){4-5}
\cmidrule(lr){6-7}
\cmidrule(lr){8-8}

& \makecell{Seen\\easy / medium / hard}
& \makecell{Unseen\\easy / medium / hard}
& \makecell{Seen\\easy / medium / hard}
& \makecell{Unseen\\easy / medium / hard}
& \makecell{Seen\\easy / medium / hard}
& \makecell{Unseen\\easy / medium / hard}
& \makecell{Unseen\\easy / medium / hard}\\
\hline
\textsc{CLS$_M$} & 70.4 / 43.7 / 41.6 & 91.7 / 40.7 / 40.5 & 38.9 / 40.1 / 40.0 & 28.1 / 34.5 / 4.3 & 44.8 / - / 51.9 & 37.1 / - / 48.3 & 94.3 / 96.2 / 96.7 \\ 
\hline
\textsc{CLS$_P$} & 100 / 99.2 / 100 & 41.7 / 83.1 / 100 & 100 / 100 / 100 & 81.2 / 83.9 / 91.3 & 100 / - / 99.7 & 74.3 / - / 93.1 & 76.1 / 62.2 / 81.3  \\
\hline
\end{tabular}
\end{adjustbox}
\caption{Performance of infeasible question detection for \textsc{CLS$_M$} and \textsc{CLS$_P$} by different hardness levels in feasible questions. The scores indicate the ratios of feasible questions not being filtered.}
\label{tab:infeasible_detection_feasible}
\end{table*}

The results in Table~\ref{tab:infeasible_detection_feasible} indicate that (1) the performances do not necessarily correlate with the hardness level, as the levels are assigned based on the query difficulty~\textbf{[P1]}, not on the question in natural language. (2) In most cases, seen questions tend to work better than unseen questions because similar questions were either trained or seen as demonstrations by the model.

\begin{table*}[h!]
\centering
\renewcommand{\arraystretch}{1.5}
\begin{adjustbox}{width=\textwidth,center}
\begin{tabular}{ccccccccccccccccccccc}
\hline

\multirow{2.5}{*}{}
 & \multicolumn{5}{c}{ATIS} & \multicolumn{5}{c}{Advising} & \multicolumn{5}{c}{EHRSQL} & \multicolumn{5}{c}{Spider}\\

\cmidrule(lr){2-6}
\cmidrule(lr){7-11}
\cmidrule(lr){12-16}
\cmidrule(lr){17-21}

& \makecell{surf} 
& \makecell{rel} 
& \makecell{unrel} 
& \makecell{non-sql}  
& \makecell{ext} 
& \makecell{surf} 
& \makecell{rel} 
& \makecell{unrel} 
& \makecell{non-sql}  
& \makecell{ext} 
& \makecell{surf}
& \makecell{rel} 
& \makecell{unrel} 
& \makecell{non-sql}  
& \makecell{ext} 
& \makecell{surf}
& \makecell{rel}
& \makecell{unrel} 
& \makecell{non-sql}  
& \makecell{ext} \\

\hline
\textsc{CLS$_M$} & 61.3 & 77.6 & 89.4 & 67.7 & 98.9 & 46.1 & 62.6 & 84.5 & 47.3 & 99.0 & 55.1 & 90.5 & 98.4 & 71.7 & 98.9 & 38.0 & 96.0 & 94.0 & 26.0 & 96.1  \\ 
\hline
\textsc{CLS$_P$} & 95.7 & 99.0 & 100 & 99.0 & 100 & 96.1 & 92.5 & 99.1 & 99.1 & 100 & 78.4 & 72.5 & 98.4 & 94.2 & 99.5 & 84.0 & 97.0 & 98.0 & 91.0 & 99.2  \\
\hline
\end{tabular}
\end{adjustbox}
\caption{Performance of infeasible question detection for \textsc{CLS$_M$} and \textsc{CLS$_P$} by five question types used for infeasible question annotation. The scores indicate the ratios of infeasible questions being filtered.}
\label{tab:infeasible_detection_infeasible}
\end{table*}

The results in Table~\ref{tab:infeasible_detection_infeasible} indicate that (1) schema incompatibility attacks, in order of difficulty, are \texttt{surface} > \texttt{unrelated} > \texttt{related}; (2) \textsc{CLS$_P$} performs well, especially in \texttt{non-sql}; and (3) \texttt{ext-know} is generally well filtered out by both methods.

\subsection{Performance of SQL Generation for Pipeline-based Methods}
\begin{table*}[h!]
\centering
\renewcommand{\arraystretch}{1.8}
\begin{adjustbox}{width=\textwidth,center}
\begin{tabular}{ccccccccccccc}
\hline

\multirow{2.5}{*}{}
 & \multicolumn{2}{c}{ATIS} & \multicolumn{2}{c}{Advising} & \multicolumn{2}{c}{EHRSQL} & \multicolumn{1}{c}{Spider}\\

\cmidrule(lr){2-3}
\cmidrule(lr){4-5}
\cmidrule(lr){6-7}
\cmidrule(lr){8-8}

& \makecell{Seen\\easy / medium / hard}
& \makecell{Unseen\\easy / medium / hard}
& \makecell{Seen\\easy / medium / hard}
& \makecell{Unseen\\easy / medium / hard}
& \makecell{Seen\\easy / medium / hard}
& \makecell{Unseen\\easy / medium / hard}
& \makecell{Unseen\\easy / medium / hard}\\
\hline
\textsc{SQLPrompt} & 100 / 93.1 / 88.8 & 58.3 / 47.5 / 40.5  & 88.9 / 91.7 / 93.3 & 78.1 / 54.0 / 34.8  & 95.4 / - / 90.6 & 71.4 / - / 53.7  & 51.1 / 60.9 / 53.8 
 \\ 
\hline
 \textsc{DIN-SQL}  & 70.4 / 25.5 / 15.7 & 25.0 / 40.7 / 7.1  & 61.1 / 62.8 / 23.3 & 62.5 / 60.9 / 34.8  & 73.6 / - / 29.2 & 60.0 / - / 31.0 & 82.9 / 74.4 / 51.6 
 \\
 \hline
\textsc{MAC-SQL}  & 77.8 / 45.3 / 66.3 & 0.0 / 39.0 / 45.2 & 70.4 / 45.5 / 36.7 & 62.5 / 46.0 / 4.3 & 70.1 / - / 40.6 & 88.6 / - / 37.4  & 77.9 / 80.1 / 49.5
\\
\hline
\textsc{T5} & 92.6 / 72.9 / 57.3 & 0.0 / 22.0 / 14.3 & 90.7 / 92.4 / 83.3 & 9.4 / 31.0 / 8.7  & 100 / - / 99.2 & 22.9 / - / 55.2  & 74.6 / 21.2 / 28.6
\\
\hline
\end{tabular}
\end{adjustbox}
\caption{Performance of SQL generators on feasible questions only, without any abstention mechanisms.}
\label{tab:sql_generation_performance}
\end{table*}
The results in Table~\ref{tab:sql_generation_performance} indicate that (1) the performances generally correlate with the hardness level, suggesting the soundness of the criteria of hardness we set. However, \textsc{MAC-SQL} and \textsc{T5} are unable to generate any correct SQL queries for unseen questions on ATIS, especially the easy samples. (2) In most cases, seen questions tend to work better than unseen questions because similar questions and SQL pairs are either trained or seen as demonstrations by the model.

\clearpage
\subsection{Performance of Error Detection for Pipeline-based Methods}

\noindent Table~\ref{tab:final_abstention_feasible} shows the performance of each method according to these scoring rules: 1) +1 for correct SQL generation without abstention, 2) +1 for incorrect SQL generation with abstention, and 3) +0 for all other cases. This scoring focuses on measuring desirable abstention scenarios for feasible questions. Table~\ref{tab:final_abstention_infeasible} shows the performance of each method according to these scoring rules: 1) +1 for abstention and 2) +0 otherwise. This scoring is designed to evaluate the methods' ability to correctly abstain from answering infeasible questions.

\begin{table*}[h!]
\centering
\renewcommand{\arraystretch}{1.8}
\begin{adjustbox}{width=\textwidth,center}
\begin{tabular}{ccccccccccccc}
\hline

\multirow{2.5}{*}{}
 & \multicolumn{2}{c}{ATIS} & \multicolumn{2}{c}{Advising} & \multicolumn{2}{c}{EHRSQL} & \multicolumn{1}{c}{Spider}\\

\cmidrule(lr){2-3}
\cmidrule(lr){4-5}
\cmidrule(lr){6-7}
\cmidrule(lr){8-8}

& \makecell{Seen\\easy / medium / hard}
& \makecell{Unseen\\easy / medium / hard}
& \makecell{Seen\\easy / medium / hard}
& \makecell{Unseen\\easy / medium / hard}
& \makecell{Seen\\easy / medium / hard}
& \makecell{Unseen\\easy / medium / hard}
& \makecell{Unseen\\easy / medium / hard}\\
\hline
\textsc{T5 → Error$_M$} & 100 / 93.5 / 92.1 & 66.7 / 61.0 / 50.0 & 100 / 94.2 / 100 & 28.1 / 46.0 / 43.5 & 100 / - / 99.2 & 37.1 / - / 66.5  & 88.6 / 81.4 / 75.8 
 \\ 
\hline
\textsc{SQLPrompt → Error$_P$} & 100 / 100 / 100 & 91.7 / 96.6 / 100 & 96.3 / 99.3 / 100 & 90.6 / 93.1 / 100 & 100 / - / 99.7 & 91.4 / - / 95.6 & 66.8 / 76.9 / 71.4 
 \\
\hline
\end{tabular}
\end{adjustbox}
\caption{Performance of error detection on feasible questions.}
\label{tab:final_abstention_feasible}
\end{table*}

We found that (1) the performances do not necessarily correlate with the hardness level for \textsc{T5 → Error$_M$}, but the performance is generally higher as the hardness level increases for \textsc{SQLPrompt → Error$_P$}, and (2) the performances of seen questions were higher than those of the unseen counterparts for each dataset.

\begin{table*}[h!]
\centering
\renewcommand{\arraystretch}{1.8}
\begin{adjustbox}{width=\textwidth,center}
\begin{tabular}{ccccccccccccccccccccc}
\hline

\multirow{2.5}{*}{}
 & \multicolumn{5}{c}{ATIS} & \multicolumn{5}{c}{Advising} & \multicolumn{5}{c}{EHRSQL} & \multicolumn{5}{c}{Spider}\\

\cmidrule(lr){2-6}
\cmidrule(lr){7-11}
\cmidrule(lr){12-16}
\cmidrule(lr){17-21}

& \makecell{surf} 
& \makecell{rel} 
& \makecell{unrel} 
& \makecell{non-sql}  
& \makecell{ext} 
& \makecell{surf} 
& \makecell{rel} 
& \makecell{unrel} 
& \makecell{non-sql}  
& \makecell{ext} 
& \makecell{surf} 
& \makecell{rel} 
& \makecell{unrel} 
& \makecell{non-sql}  
& \makecell{ext} 
& \makecell{surf} 
& \makecell{rel} 
& \makecell{unrel} 
& \makecell{non-sql}  
& \makecell{ext} \\

\hline
\textsc{T5 → Error$_M$} & 81.7 & 86.7 & 90.4 & 86.9 & 96.7 & 61.8 & 63.6 & 86.4 & 76.4 & 97.1 & 74.6 & 56.6 & 94.7 & 72.8 & 61.0 & 63.0 & 75.0 & 92.0 & 68.0 & 94.5 
 \\ 
\hline
\textsc{SQLPrompt → Error$_P$} & 89.2 & 92.9 & 71.3 & 59.6 & 46.7 & 92.2 & 84.1 & 71.8 & 54.5 & 15.4 & 84.9 & 92.6 & 86.1 & 39.3 & 28.0 & 62.0 & 57.0 & 53.0 & 43.0 & 54.3  \\
\hline
\end{tabular}
\end{adjustbox}
\caption{Performance of error detection on infeasible questions.}
\label{tab:final_abstention_infeasible}
\end{table*}
The results in Table~\ref{tab:final_abstention_infeasible} indicate that (1) schema incompatibility attacks, in order of difficulty, are \texttt{surface} > \texttt{unrelated} > \texttt{related}; (2) \textsc{T5 → Error$_M$} detects \texttt{non-sql} and \texttt{ext-know} much better than \textsc{SQLPrompt → Error$_P$}.


\clearpage
\section{Data Preprocessing}
\label{appendix:data_preprocessing}
In analyzing the ATIS and Advising datasets\footnote{We utilized the improved version of the original datasets, where the questions are grouped together in terms of the same SQL structures~\citep{advising}.}, we observed several annotation issues: query duplication, limited paraphrase diversity per SQL query, null results, and data inconsistencies in both NL and SQL data. In this section, we outline the preprocessing steps taken to address these issues.

\paragraph{Query duplication} Significant query duplication was found in both ATIS and Advising, where questions conveying the same meaning are matched with differently labeled SQL structures~\citep{advising}. While these are not critical in terms of data quality, this may result in misleading outcomes when evaluating model performance on `seen SQL' and `unseen SQL' settings after data splitting. To address this, we identified NL paraphrases sharing identical placeholders (\textit{e.g.}, `I need a flight from city\_name1 to city\_name0' and `Which airline provides service in city\_name0 and city\_name1' both use city\_name0 and city\_name1 as placeholders.) and then assessed their semantic equivalence. For any SQL structures with identical semantics, we kept the one with more NL paraphrases and merged the others into it.


\paragraph{Limited paraphrase diversity}  Our findings indicate that over 90\% of SQL queries in ATIS have fewer than 10 paraphrases per SQL query (\textit{e.g.}, `I need a flight from city\_name1 to city\_name0' and `Give me a list of flights going from city\_name1 to city\_name0' are two different paraphrases for the same SQL structure). Additionally, both ATIS and Advising exhibit limited diversity and quality in NL questions. To improve question quality, we used ChatGPT to generate 10 paraphrases for each SQL query, followed by manual review to ensure they reflect the intended SQL meaning.

During our review, we specifically focused on avoiding inconsistencies. We noticed that the original questions in ATIS and Advising often contain mismatches between the columns targeted by the SQL queries and those referenced in the NL questions. For example, there were multiple instances where an NL question referred to people's names, but the corresponding SQL query selected their IDs instead of their names \textit{e.g.}, INSTRUCTOR.INSTRUCTOR\_ID rather than INSTRUCTOR.INSTRUCTOR\_NAME). This could result in potential false positives in execution accuracy.

\paragraph{Null results} We refer to null results as cases where ground-truth SQLs return `Null' or `[]' as a result of execution, which may cause false positives in text-to-SQL model evaluation. We observed numerous data samples in Advising (and some in ATIS) fitting this scenario. Consequently, we manually corrected these cases in both ATIS and Advising by sampling a `valid' combination of placeholder values for the questions and SQLs. The word `combination' is used here because questions often contain multiple placeholders, which need their realized values to coexist for successful execution without null results.

\paragraph{Inconsistent text-to-SQL assumptions} ATIS and Advising contains questions with relative time expressions lacking a clear current time reference (\textit{e.g.}, I need a flight from city\_name1 to city\_name0 `today'). Moreover, the current time, inferred from NL questions and evident in SQL queries, varies across samples. Additionally, Advising contains SQL queries that assume a specific student without explicitly mentioning that in NL questions (\textit{e.g.}, Can `I' take department0 number0 in semester0 year0?). To resolve these inconsistencies, we converted relative time expressions to their absolute-time equivalents (\textit{e.g.}, `today' => `on month\_number0 / day\_number0 / year0'), and specified the student's identity when needed.

\clearpage
\section{URL to GitHub and Croissant Metadata Record}

\begin{itemize}
     \item Github: \url{https://github.com/glee4810/TrustSQL}
    \item Croissant Metadata Record: \url{https://huggingface.co/api/datasets/glee4810/TrustSQL}
\end{itemize}

\section{Author Statement}
The authors bear all responsibility in case of violation of rights, etc. associated with the TrustSQL dataset.


\end{document}